\documentclass[11pt]{article}

\usepackage[final]{acl}

\usepackage{times}
\usepackage{latexsym}

\usepackage[T1]{fontenc}

\usepackage[utf8]{inputenc}

\usepackage{microtype}

\usepackage{inconsolata}
\usepackage{booktabs}   
\usepackage{multirow}   
\usepackage{tabularx}   
\usepackage{array}      
\usepackage{amsmath}
\usepackage{amssymb}
\usepackage{colortbl}
\usepackage{graphicx}
\usepackage{siunitx}
\usepackage{color}
\usepackage{xspace}

\title{QuantumQA: Enhancing Scientific Reasoning via Physics-Consistent Dataset and Verification-Aware Reinforcement Learning}

\author{
 \textbf{Songxin Qu\textsuperscript{1,2}},
 \textbf{Tai-Ping Sun\textsuperscript{3}},
  \textbf{Yun-Jie Wang\textsuperscript{1}},
  \textbf{Huan-Yu Liu\textsuperscript{2}},
 \textbf{Cheng Xue\textsuperscript{2}},\\
 \textbf{Xiao-Fan Xu \textsuperscript{3}},
  \textbf{Han Fang \textsuperscript{5}},
 \textbf{Yang Yang\textsuperscript{4}},
  \textbf{Yu-Chun Wu\textsuperscript{3}},
  \textbf{Guo-Ping Guo\textsuperscript{3}},
  \textbf{Zhao-Yun Chen\textsuperscript{2}}
\\
 \textsuperscript{1}Institute of Advanced Technology, University of Science and Technology of China
 \\
 \textsuperscript{2}Institute of Artificial Intelligence, Hefei Comprehensive National Science Center
 \\
 \textsuperscript{3}School of Physics, University of Science and Technology of China
 \\
 \textsuperscript{4}School of Electronics and Information Engineering, Anhui University\\
  \textsuperscript{5}School of Computing, National University of Singapore\\
 \small{
   \textbf{Correspondence:} \href{gpguo@ustc.edu.cn}{gpguo@ustc.edu.cn}, \href{chenzhaoyun@iai.ustc.edu.cn}{chenzhaoyun@iai.ustc.edu.cn}
 }
}

\begin{document}
\maketitle

\begin{abstract}
Large language models (LLMs) show strong capabilities in general reasoning but typically lack reliability in scientific domains like quantum mechanics, which demand strict adherence to physical constraints. This limitation arises from the scarcity of verifiable training resources and the inadequacy of coarse feedback signals in standard alignment paradigms. To address the data challenge, we introduce \textsc{QuantumQA}, a large-scale dataset constructed via a task-adaptive strategy and a hybrid verification protocol that combines deterministic solvers with semantic auditing to guarantee scientific rigor. Building on this foundation, we propose the verification-aware reward model (VRM) tailored for Reinforcement Learning with Verifiable Rewards (RLVR), which employs an adaptive reward fusion (ARF) mechanism to dynamically integrate deterministic signals from a scientific execution suite (SES) with multidimensional semantic evaluations for precise supervision. Experimental results demonstrate that our method consistently outperforms baselines and general-purpose preference models. Notably, our optimized 8B model achieves performance competitive with proprietary models, validating that incorporating verifiable, rule-based feedback into the reinforcement learning loop offers a parameter-efficient alternative to pure scaling. 
\end{abstract}

\section{Introduction}

Large language models (LLMs) have shown strong capabilities in general reasoning and mathematical problem solving~\citep{wei2022chain,shao2025deepseekmath,glazer2025frontiermathbenchmarkevaluatingadvanced}. 
However, their reliability in scientific domains that require strict adherence to axioms and physical constraints remains inconsistent~\citep{wangscibench,taylor2022galacticalargelanguagemodel}. 
This gap arises less from an absence of general reasoning ability than from limited access to domain-specific constraints and systematic verification that guide scientific reasoning beyond purely mathematical validity~\citep{yao2023react}. 
Quantum mechanics, with a compact axiomatic foundation and well-defined constraint structure across common approximation regimes~\citep{von2018mathematical,nielsen2010quantum}, therefore offers a stringent and controlled testbed for studying constrained scientific reasoning in LLMs~\citep{guo2025quanbenchbenchmarkingquantumcode, minami2025quantumbenchbenchmarkquantumproblem}.

Applying LLMs to this rigorous domain faces two primary challenges. The first is the scarcity of high-quality, verifiable training resources. 
Existing resources typically bifurcate into two extremes: they are either small-scale, multiple-choice benchmarks inadequate for training complex reasoning~\citep{minami2025quantumbenchbenchmarkquantumproblem}, or large-scale synthetic corpora that lack physical verification mechanisms~\citep{kashani2024quantumllminstruct500kllminstructiontuning}.
While recent datasets address quantum algorithm implementation and code generation~\citep{yang2025qcircuitbench, vishwakarma2024qiskithumanevalevaluationbenchmark, paltenghi2024surveytestinganalysisquantum}, and low-level quantum circuit design and compilation tasks~\citep{foder2024reinforcementlearningvariationalquantum, fu2025qagentllmbasedmultiagentautonomous, zhang2024scalablequantumdynamicscompilation}, they primarily focus on algorithmic implementation and circuit-level tasks.
The lack of scalable, process-supervised data constrains models' mathematical reasoning capabilities in quantum theory~\citep{liu-etal-2024-mathbench, vishwakarma2024qiskithumanevalevaluationbenchmark}.

The second critical challenge lies in the limitations of existing alignment methods within scientific domains.
Standard Reinforcement Learning from Human Feedback (RLHF)~\citep{ouyang2022training} suffers from reward over-optimization~\citep{gao2023scaling, skalse2022defining, taylor2022galacticalargelanguagemodel, miao2025informationtheoreticrewardmodelingstable} and often yielding plausible-sounding hallucinations that violate physical constraints~\citep{chua2025learning, taylor2022galacticalargelanguagemodel}.
Although Process Reward Models (PRMs) provide dense supervision for formal domains like math~\citep{setlur2025rewarding, yang2025errorprocessrewardmodels} and coding~\citep{li-etal-2025-codeprm, zhang2025dreamprmcodefunctionasstepprocessreward} by leveraging step-wise verification, such verifiers are often scarce in scientific settings. 
While RLVR incorporates execution signals~\citep{gou2024tora, gou2024critic}, it typically relies on sparse, outcome-based supervision~\citep{lightman2023let, uesato2022solving}, which is insufficient for complex scientific problems. 
Furthermore, existing multi-objective frameworks prioritize balancing general-purpose human preferences, such as helpfulness and safety~\citep{wang-etal-2024-interpretable, bai2022training, dai2024safe}, rather than enforcing verifiable correctness or physical consistency. 

To systematically address the challenges of data scarcity and reasoning hallucination, we build upon the Seed-Evolve paradigm~\citep{wang-etal-2023-self-instruct, zeng-etal-2024-automatic} to expand topical breadth, while introducing two rigor-focused enhancements to ensure scientific validity. First, we implement a task-adaptive data construction strategy, which effectively mitigates hallucination by tailoring response structures to task complexity~\citep{wang-etal-2023-plan}.
Specifically, this approach enforces conciseness for straightforward tasks while mandating detailed Chain-of-Thought (CoT) derivations for complex reasoning~\citep{magister-etal-2023-teaching,wei2022chain}. 
Second, to guarantee scientific rigor in synthetic data, we devise a hybrid verification protocol that integrates deterministic verification tools via our scientific execution suite (SES), a deterministic toolkit consisting of automated verification scripts and symbolic solvers~\citep{meurer2017sympy}, with semantic auditing~\citep{zheng2023judging}, culminating in a Human-in-the-Loop (HITL) review process to ensure the quality of dataset~\citep{cobbe2021trainingverifierssolvemath, ouyang2022training}.
By strictly enforcing physical constraints, this mechanism effectively purges plausible yet fallacious derivations that evade standard self-evaluation~\citep{JiHallucination2023}.

To address the challenge of feedback reliability, we introduce the verification-aware reward model (VRM), a unified reward model tailored for RLVR.
VRM combines deterministic verification signals from the SES~\citep{lightman2023let} with semantic evaluation provided by an LLM-as-a-Judge framework~\citep{zheng2023judging}. 
Specifically, the semantic evaluation encompasses three key dimensions: Mathematical Correctness (\textit{Corr}), Physical Consistency (\textit{Phys}), and Instruction Following (\textit{Inst}). 
Additionally, we employ an adaptive reward fusion (ARF) mechanism that dynamically modulates supervision strength based on verifiability~\citep{uesato2022solving}.

This dynamic calibration mitigates the reward sparsity often found in rigid execution environments, effectively bridging the gap between general preference optimization~\citep{schulman2017proximal} and the rigorous supervision required for complex scientific reasoning~\citep{schick2023toolformer}.

We conduct comprehensive evaluations on the test subset from \textsc{QuantumQA} and external scientific benchmarks (\textsc{SuperGPQA}~\citep{pteam2025supergpqascalingllmevaluation} and \textsc{Physics})~\citep{feng-etal-2025-physics}.
The experimental results demonstrate that our method consistently outperforms supervised fine-tuning (SFT)~\citep{zhang2023instruction,dong-etal-2024-abilities} across diverse open-source backbones. 
Notably, our VRM-optimized 8B model achieves performance competitive with proprietary systems like ChatGPT and DeepSeek-R1~\citep{Guo_2025} in certain metrics. 
These findings suggest that incorporating verifiable, rule-based feedback into the reinforcement learning loop offers a parameter-efficient alternative to pure scaling.
To elucidate the source of these gains, we perform fine-grained error analysis and component ablations. We observe that generic preference models struggle to distinguish subtle calculation errors from plausible hallucinations, whereas our verification-aware signals effectively reduce logical violations and mitigate the length-bias often associated with RLHF.
Furthermore, ablation studies indicate that our ARF plays a crucial role in stabilizing optimization.
These results suggest that fine-grained, multi-dimensional supervision significantly enhances the reliability of automated scientific reasoning.

In summary, our contributions are as follows:

1. We construct \textsc{QuantumQA}, a large-scale dataset of 92,749 samples for verifiable scientific reasoning. By leveraging a task-adaptive data construction pipeline and a hybrid verification protocol, we guarantee scientific rigor and high data validity. \textsc{QuantumQA} will be released upon publication.

2. We propose the VRM, which combines verification feedback and multidimensional semantic evaluation to enable precise optimization of scientific reasoning.

3. Experiments demonstrate that our method achieves superior performance and parameter efficiency, surpassing proprietary models like ChatGPT-5. Furthermore, it significantly outperforms SFT baselines across several base models as well as generic preference models.

\section{The QuantumQA Dataset}
\label{sec:dataset}

To facilitate robust alignment and evaluation in physics-constrained reasoning, we introduce \textsc{QuantumQA}, a verified, multi-task dataset comprising 92,749 samples.
Designed to overcome the limitations of previous works, \textsc{QuantumQA} supports both SFT and Reinforcement Learning pipelines, as well as serving as a rigorous evaluation benchmark. 

\subsection{Dataset Construction and Verification}
\label{sec:pipeline}

\begin{figure}[t]
  \centering
  \includegraphics[width=\columnwidth]{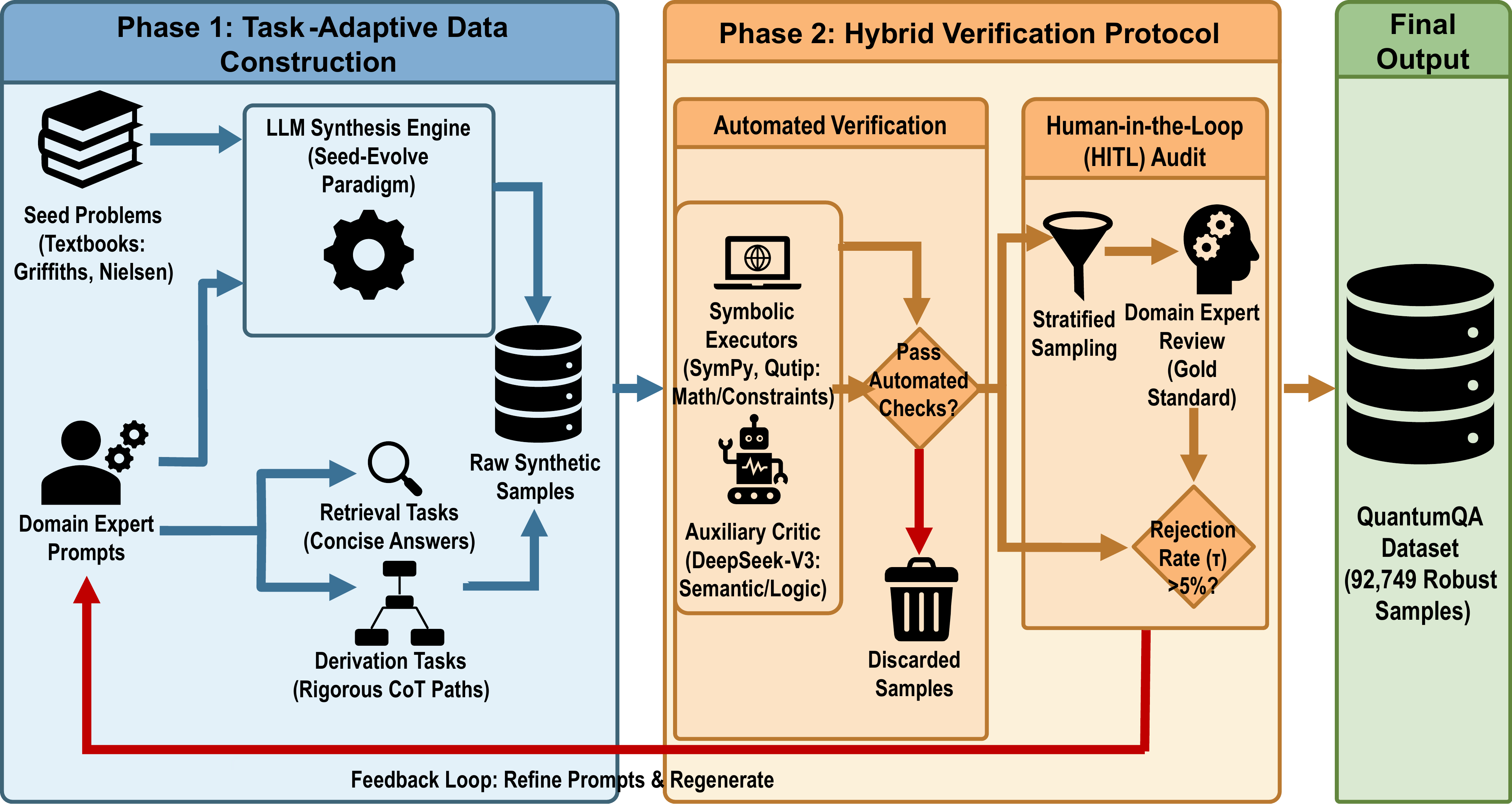}
  \caption{Dataset construction and verification pipeline.}
  \label{fig:data_construct}
\end{figure}

A core challenge in synthesizing scientific data lies in mitigating reasoning hallucination.
As shown in Fig.~\ref{fig:data_construct}, we refine the standard synthetic data generation pipeline~\citep{yu2023metamath, shao2024deepseekmath} by incorporating two rigor-focused enhancements: a task-adaptive data construction strategy and a hybrid verification protocol. This approach ensures that the dataset aligns with strict physical constraints while maintaining the diversity inherent in large-scale synthesis.
Comprehensive implementation details are provided in Appendix~\ref{sec:appendix_pipeline}.

\paragraph{Task-Adaptive Data Construction.}
We build upon the established Seed-Evolve paradigm~\citep{wang-etal-2023-self-instruct} to expand topical breadth from authoritative seed problems~\citep{griffiths2018introduction, nielsen2010quantum}. However, to address the distinct cognitive demands of different scientific tasks, we introduce a task-adaptive construction strategy. Specifically, retrieval-heavy tasks are constrained to concise answers to prevent verbose hallucinations~\citep{zhou2023lima}, whereas complex tasks are augmented with rigorous CoT paths to facilitate precise process supervision~\citep{wei2022chain, lewkowycz2022solving}.

\paragraph{Hybrid Verification Protocol.}

To guarantee scientific rigor, we implement a dual-stage validation mechanism integrating automated verification with HITL auditing. 
Specifically, we employ SES to enforce physical constraints while leveraging an independent LLM as a semantic critic~\citep{meurer2017sympy, johansson2012qutip, zheng2023judging}. 
Following this, we conduct a HITL audit where a dynamic rejection threshold ($\tau > 5\%$) triggers iterative refinement to establish a high-reliability gold standard~\citep{cobbe2021trainingverifierssolvemath}.

\subsection{Dataset Analysis}
\label{sec:analysis}

We analyze the characteristics of \textsc{QuantumQA} to demonstrate its diversity and detail the data splits tailored for our optimization.
\paragraph{Diversity and Coverage.}
Instead of relying on simple pattern matching, \textsc{QuantumQA} encompasses five distinct types of tasks, including Short Answer, Fill-in-the-Blank, True/False, Multiple Choice, and Problem Solving, where the Problem Solving task is emphasized for the requirement of multi-step derivation. 
Furthermore, the dataset spans a broad spectrum of topics, from theoretical foundations to physical implementations, while maintaining a balanced difficulty distribution to assess varying cognitive loads. 
Details are provided in Appendix~\ref{sec:appendix_stats}.

\paragraph{Data Splits.}
We stratify \textsc{QuantumQA} into train ($\approx$95\%), dev ($\approx$100), and a strictly held-out test set ($\approx$5\%, $N=4{,}675$).
Within the training split, we allocate $\approx$70\% for SFT training and construct the RLVR training set from the remaining challenging prompts, augmented with a small pool ($\approx$5\%) of high-complexity prompts mined during SFT.

\section{Verification-Aware Reward Model}
\label{sec:method}
\begin{figure*}[htbp]
  \centering
  \includegraphics[width=\textwidth]{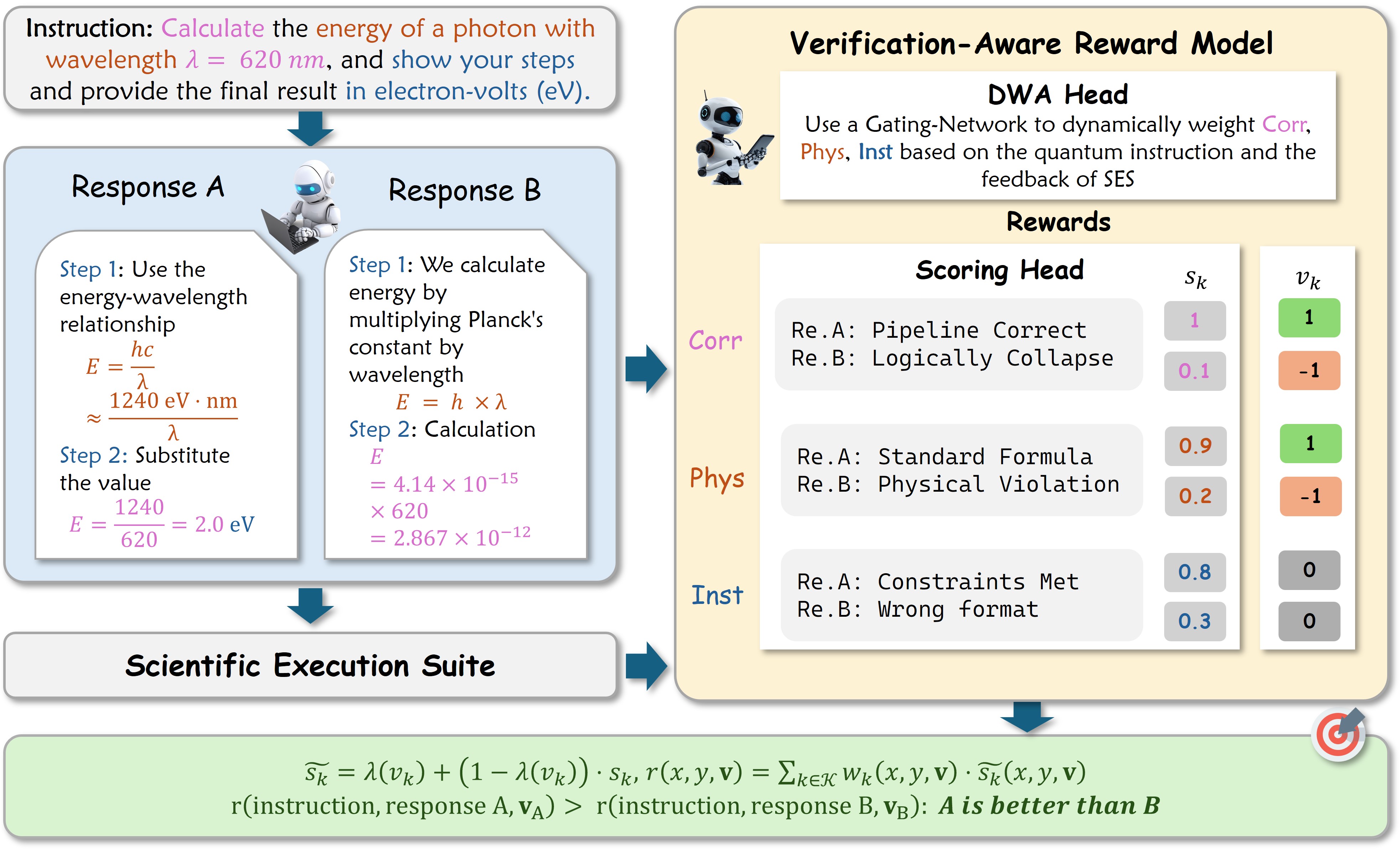}
  \caption{Overview of the verification-aware reward model.}
  \label{fig:model_arch}
\end{figure*}

In this section, we propose VRM, a verification-driven reward mechanism designed to align language model generation.
As illustrated in Fig.~\ref{fig:model_arch}, VRM integrates intrinsic semantic assessment with extrinsic tool verification.
Concretely, the model operates through three coordinated stages. 
First, the SES performs code-based validation to yield deterministic feedback vector $\mathbf{v}$. 
Simultaneously, the Scoring head evaluates the response across multiple semantic dimensions to produce semantic scores $\mathbf{s}$. 
To synthesize these heterogeneous signals, we introduce the Dynamic Weight Allocation (DWA) head, a gating network that dynamically estimates reliability weights based on the execution results. 
Finally, these components are aggregated into a robust scalar reward, thus enabling the model to effectively distinguish superior responses even in the presence of partial verification noise. 

\subsection{Verification Signals from SES}
\label{subsec:verification_signals}

Given an input context $x$ and a candidate response $y$, we evaluate the pair $(x, y)$ along a set of dimensions $\mathcal{K}$. 
To formalize the verification signals, we partition the evaluation dimensions $\mathcal{K}$ into verifiable ($\mathcal{K}_{\text{ver}} = \{\text{Corr}, \text{Phys}\}$) and purely semantic ($\mathcal{K}_{\text{sem}} = \{\text{Inst}\}$) subsets. 
Specifically, dimensions in $\mathcal{K}_{\text{ver}}$ leverage the SES to yield deterministic execution feedback, whereas $\mathcal{K}_{\text{sem}}$ relies on semantic judgments due to the absence of formal solvers.

Accordingly, the feedback $\mathbf{v}$ composed of trinary verification indicators $v_k$ for each $k \in \mathcal{K}$.
The indicator $v_k \in \{1, 0, -1\}$ captures the detailed execution status:
$$v_k = \begin{cases} 
1 & \text{constraint satisfied,} \\
-1 & \text{violation detected,} \\
0 & \text{execution unavailable.}
\end{cases}$$

Finally,  the vector $\mathbf{v}$ is concatenated with the textual representation of $(x, y)$ to serve as inputs for the VRM.

\subsection{Verification-Aware Reward Architecture}
\label{subsec:agent}

Our architecture is founded on a pre-trained Transformer backbone $\mathcal{M}$ (instantiated via Qwen3-4B) that serves as a shared encoder, where the standard causal head is replaced by two specialized output modules. Given $(x,y)$, the model first extracts a unified contextual representation:
\begin{equation}
\mathbf{h} = \mathcal{M}(x, y).
\end{equation}
Based on this shared embedding $\mathbf{h}$, we design two parallel prediction heads, structured as 3-layer MLPs with 1,024 hidden units and GeLU activations, to jointly estimate generation quality and reliability weights.

\paragraph{Scoring Head.}
First, to assess the semantic plausibility of the generation, especially when SES is inapplicable, we employ a Scoring head  to project the hidden state $\mathbf{h}$ into a semantic evaluation score:

\begin{equation}
         \mathbf{s} = \sigma(\mathbf{W}_s \mathbf{h} + \mathbf{b}_s) \in [0,1]^{|\mathcal{K}|},
\end{equation}
where $\mathbf{W}_s$ and $\mathbf{b}_s$ are learnable parameters. This score serves as an intrinsic quality estimation, providing a dense supervision signal regardless of tool usage.

\paragraph{DWA Head.}
Simultaneously, it dynamically estimates the reliability weight of each dimension. By concatenating the semantic embedding $\mathbf{h}$ with $\mathbf{v}$, this head computes a dimension-wise weight vector:
\begin{equation}
    \mathbf{w} = \sigma(\mathbf{W}_g [\mathbf{h} \,;\, \mathbf{v}] + \mathbf{b}_g) \in [0,1]^{|\mathcal{K}|},
\end{equation}
where $\mathbf{W}_g$ and $\mathbf{b}_g$ are learnable parameters.
This head learns to adjust the corresponding weights with reliable SES feedback, facilitating robust reward estimation.

\subsection{Adaptive Reward Fusion}
\label{sec:method_reward}

Given the verification indicator $\mathbf{v}$, the semantic evaluation score $\mathbf{s}$, and the reliability weights $\mathbf{w}$, the VRM produces a scalar reward used by the downstream reinforcement learning algorithm.
\paragraph{Signal Fusion.}
For each dimension $k$, we compute a fused score $\tilde{s}_k$:
\begin{equation}
    \tilde{s}_k = \lambda(v_k) + \bigl(1 - \lambda(v_k)\bigr) \cdot s_k,
\end{equation}
where $s_k$ denotes the $k$-th scalar component of $\mathbf{s}$. 
We set $\lambda(1) = 1.0$ to fully trust successful verification and $\lambda(0) = 0.0$ to fall back entirely on the semantic evaluation score when a reliable execution result is unavailable. For failed verification ($v_k = -1$), we use a small constant $0 < \lambda(-1) \ll 1$ to weaken the influence of SES, while preserving the contribution of the fine-grained semantic assessment $s_k$ to ensure the stability of optimization.


\paragraph{Reliability-Weighted Aggregation.}
The final scalar reward is obtained as a reliability-weighted sum over fused scores:
\begin{equation}
    r(x, y, \mathbf{v}) = \sum_{k \in \mathcal{K}} w_k(x,y,\mathbf{v}) \cdot \tilde{s}_k(x,y,\mathbf{v}).
\end{equation}
As a result, optimization is dominated by fused scores and reliable weights.

\subsection{Oracle-Guided Pretraining}
\label{subsec:oracle_training}

To construct the VRM, we perform supervised distillation from an ``Oracle-as-a-Judge'' pipeline.
Initially, we collect base tuples consisting of instructions, responses, and tool execution traces, denoted as $(x, y, \mathbf{v})$.
To mitigate self-preference bias, we then employ an ensemble of oracles to evaluate these tuples, generating soft quality ratings $\mathbf{s}^*$ and dimension importance weights $\mathbf{w}^*$. 
By combining these elements, we construct the comprehensive dataset $\mathcal{D} = \{(x, y, \mathbf{v}, \mathbf{s}^*, \mathbf{w}^*)\}$. 
Crucially, our oracle models are strictly disjoint from downstream baselines and judges, eliminating information leakage and ensuring the VRM learns generalized verification over model-specific biases. 
The VRM is then trained to jointly optimize score regression and weight prediction via a multi-task objective:

\begin{equation}
    \mathcal{L} = \mathbb{E}_{(x,y,\mathbf{v}) \sim \mathcal{D}} \left[ \|\mathbf{w} - \mathbf{w}^*\|_2^2 + \beta \,\|\mathbf{s} - \mathbf{s}^*\|_2^2 \right],
\end{equation}
where $\beta$ is a balancing coefficient. 
This alignment phase enables the VRM to serve as an efficient, standalone reward model for subsequent reinforcement learning, eliminating the need for expensive oracle queries.

\section{Experiments}
\label{sec:Experiments}

This section evaluates our work across multiple aspects. We detail the experimental setup in \S~\ref{subsec:Experimental Setup} and present the main comparative results in \S~\ref{subsec:exp_results}. To demonstrate the robustness of our approach, we then assess its generalizability to alternative reinforcement learning algorithms in \S~\ref{sec:grpo_generalization} and out-of-distribution scientific benchmarks in \S~\ref{subsec:generalization}. Furthermore, we provide a fine-grained analysis of model behaviors, specifically regarding verbosity (\S~\ref{subsec:analysis_verbosity}) and error distribution (\S~\ref{subsec:error_analysis}). Finally, we conduct ablation studies in \S~\ref{subsec:ablation} to isolate component contributions.

\begin{table*}[t]
  \centering
  \resizebox{\textwidth}{!}{ 
  \begin{tabular}{llccccc}
    \toprule
    \multirow{2}{*}{\textbf{Models}} & \multirow{2}{*}{\textbf{Method}} & \textbf{Fill-in-the-blank} & \textbf{True/False} & \textbf{Multiple Choice} & \textbf{Problem Solving} & \textbf{Short Answer} \\
    \cmidrule(lr){3-3} \cmidrule(lr){4-4} \cmidrule(lr){5-5} \cmidrule(lr){6-6} \cmidrule(lr){7-7}
     & & Exact Match $\uparrow$ 
    & Accuracy $\uparrow$ 
    & Accuracy $\uparrow$ 
    & $\mathrm{ACC}_{\mathrm{U}}$ $\uparrow$ 
    & $\mathrm{ACC}_{\mathrm{U}}$ $\uparrow$ \\
    \midrule
    \multirow{3}{*}{Qwen3-8B} & Baseline & 0.289 & 0.841 & 0.895 & 0.512 & 0.792 \\
     & SFT & 0.500 & 0.899 & 0.967 & 0.621 & 0.817 \\
     & RLVR (ours) & \textbf{0.578} & \textbf{0.912} & 0.978 & \textbf{0.680} & \textbf{0.897} \\
     & RLVR (GRPO) & 0.568 & 0.904 & \underline{\textbf{0.979}} & 0.655 & 0.875 \\
    \midrule
    \multirow{3}{*}{Meta-Llama-3.1-8B-Instruct} & Baseline & 0.318 & 0.792 & 0.851 & 0.374 & 0.599 \\
     & SFT & 0.494 & 0.876 & 0.959 & 0.517 & 0.799 \\
     & RLVR (ours) & \textbf{0.564} & \textbf{0.908} & \textbf{0.971} & \textbf{0.614} & \textbf{0.877} \\
     \midrule
    \multirow{3}{*}{DeepSeek-R1-0528-Qwen3-8B} & Baseline & 0.189 & 0.829 & 0.807 & 0.561 & 0.807 \\
     & SFT & 0.442 & 0.881 & 0.948 & 0.579 & 0.821 \\
      & RLVR (ours) & \textbf{0.487} & \textbf{0.888} & \textbf{0.960} & \textbf{0.608} & \textbf{0.850} \\
    
    \midrule 
    Human-Expert & N/A & \underline{0.720} & \underline{0.950} & 0.975 & 0.700 & 0.925 \\
    ChatGPT-5 & Baseline & 0.310 & 0.894 & 0.910 & 0.642 & 0.829 \\
    DeepSeek-R1-0528 & Baseline & 0.183 & 0.847 & 0.866 & 0.665 & 0.952 \\
    GLM-4.6 & Baseline & 0.361 & 0.839 & 0.947 & 0.708 & 0.952 \\
    Qwen3-Max & Baseline & 0.491 & 0.903 & 0.957 & \underline{0.804} & \underline{0.977} \\
    Kimi-K2-Thinking & Baseline & 0.323 & 0.815 & 0.862 & 0.689 & 0.966 \\
    Qwen3-235B-A22B-Instruct & Baseline & 0.338 & 0.875 & 0.932 & 0.847 & 0.952 \\
    \bottomrule
  \end{tabular}
  }
  \caption{Zero-shot performance comparison on QuantumQA using greedy decoding. Note: \textbf{Bold} font indicates the best performance within each model group. \underline{Underline} denotes the overall best result in each column.}
  \label{tab:model_comparison_final}
\end{table*}

\subsection{Experimental Setup}
\label{subsec:Experimental Setup}
\paragraph{Backbones and Baselines.}
We validate our framework across a spectrum of open-source backbones, primarily focusing on the Qwen3 family~\citep{qwen3technicalreport} and the LLaMA3 series~\citep{grattafiori2024llama3herdmodels}, as well as DeepSeek-R1-0528-Qwen3-8B. To thoroughly benchmark the effectiveness of our proposed method, we compare the following experimental settings:
(1) \textbf{Baseline}: The unadapted open-source models;
(2) \textbf{SFT}: Supervised fine-tuning on \textsc{QuantumQA};
(3) \textbf{Skywork RM}: Proximal Policy Optimization (PPO)~\citep{schulman2017proximal} training guided by Skywork-Reward~\citep{liu2025skywork}, a state-of-the-art general-purpose reward model, serving as a baseline for scalar-reward RL;
(4) \textbf{RLVR (Ours)}: PPO training driven by our VRM, serving as our primary optimization strategy;
(5) \textbf{RLVR (GRPO)}: Group Relative Policy Optimization (GRPO) \citep{shao2024deepseekmath} training guided by our VRM, evaluated to demonstrate the algorithm-agnostic generalizability of our approach.
Additionally, we include frontier proprietary models (e.g., ChatGPT-5, Qwen3-Max) and human expert performance (evaluated on a representative stratified sample of 400 instances) as upper-bound references to contextualize the reasoning gap.

\paragraph{Benchmarks and Datasets.}

We conduct our primary evaluation on the held-out test split of \textsc{QuantumQA}. To further verify the robustness of our approach, we extend our evaluation to the quantum mechanics subsets of \textsc{SuperGPQA}~\citep{pteam2025supergpqascalingllmevaluation} and \textsc{Physics}~\citep{feng-etal-2025-physics}.
We employ these subsets as necessary alternatives to other domain-specific benchmarks, such as \textsc{QuantumBench}~\citep{minami2025quantumbenchbenchmarkquantumproblem} and \textsc{QuantumLLMInstruct}~\citep{kashani2024quantumllminstruct500kllminstructiontuning}, which are excluded from this study due to access restrictions and data incompleteness, respectively.

\paragraph{Evaluation Metrics.}
\label{par:Evaluation Metrics}
We adopt task-specific metrics tailored to the output format: \textit{Accuracy} for Multiple-Choice/True-False tasks and \textit{Exact Match} for Fill-in-the-blank tasks. For open-ended Problem Solving and Short Answer tasks, we utilize a unified score, $\mathrm{ACC}_{\mathrm{U}}$, derived via an LLM-as-a-judge protocol~\citep{liu-etal-2023-g,vertselHybridLLMRulebased2024,zheng2023judging,liLLMsasjudgesComprehensiveSurvey2024}. Specifically, to mitigate the self-preference bias inherent in model-based evaluation~\citep{zheng2023judging, wang-etal-2024-large-language-models-fair}, we employ Qwen3-Max to score responses on a normalized scale of $[0,1]$ based on reasoning rigor and correctness. 
To ensure evaluation reliability, we validated this automated protocol against human expert annotations on a subset of the data, observing a high Spearman correlation. 
For other standard benchmarks, we follow their official evaluation protocols, normalizing results to $[0,1]$ for consistency.
Detailed validation results are provided in Appendix~\ref{app:judge_validation}.

\subsection{Main Results}
\label{subsec:exp_results}

Table~\ref{tab:model_comparison_final} presents the primary evaluation results. 
We conduct a comprehensive analysis to evaluate the overall efficacy and practical advantages of our proposed approach.

\begin{figure}[t]
    \centering
    \includegraphics[width=0.9\linewidth]{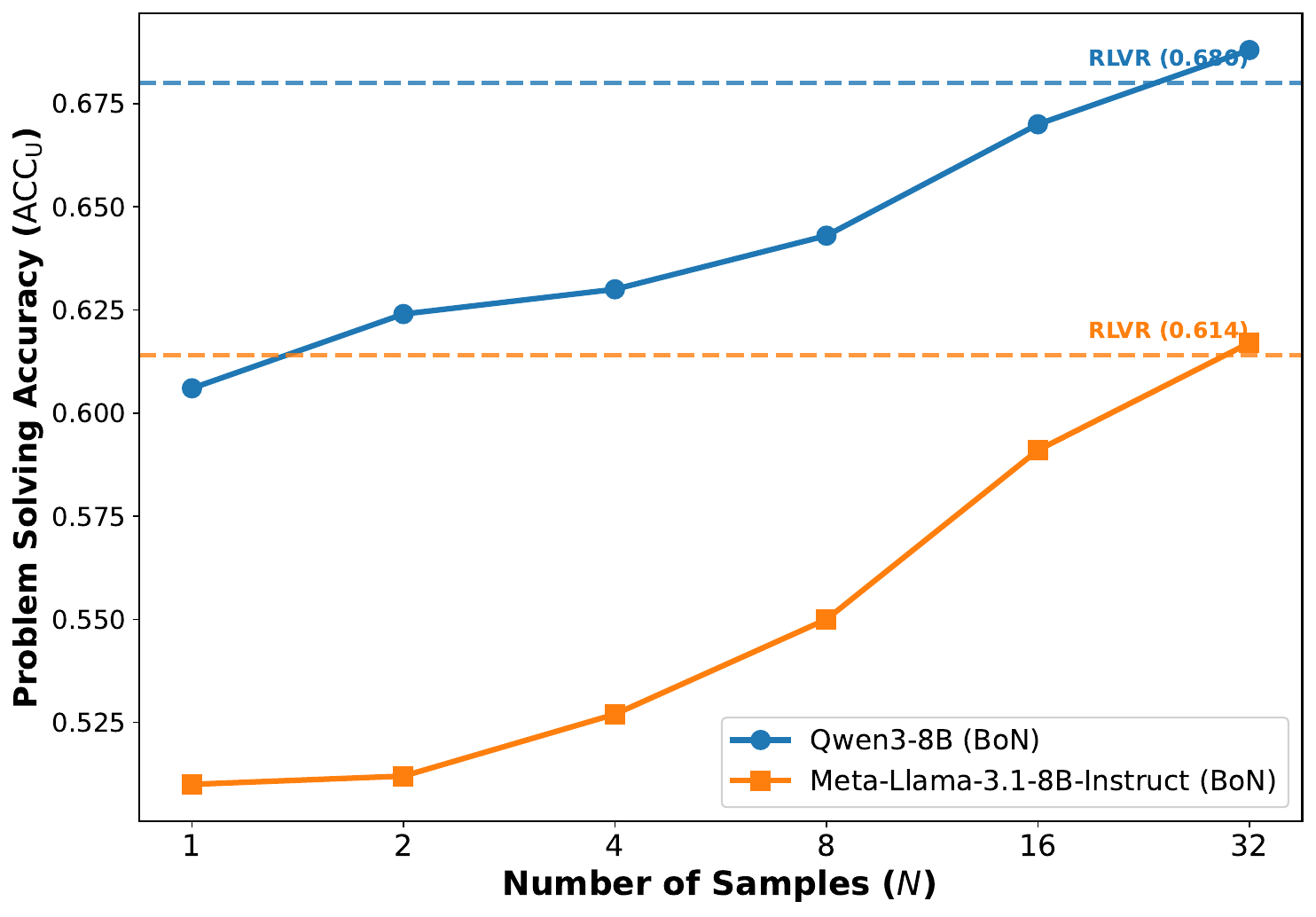}
    \caption{Best-of-$N$ performance scaling.
    }
    \label{fig:bon_scaling}
\end{figure}
\paragraph{Effectiveness of Supervised Fine-Tuning.}
Comparing SFT against base models, we observe consistent performance gains across all metrics. This confirms that domain-specific supervision is essential for injecting quantum knowledge. However, the reliance on static ground-truth labels limits SFT's generalization in complex reasoning, motivating the integration of verification-based reinforcement learning.

\paragraph{Impact of Verification-Aware RL.}
Applying RLVR yields significant improvements over SFT baselines. For Qwen3-8B, our method improves Problem Solving accuracy from $0.621$ to $0.680$ and Short Answer accuracy from $0.817$ to $0.897$. 
Similar trends are observed with Meta-Llama-3.1-8B-Instruct. 
These gains indicate that our VRM effectively steers the model toward rigorous reasoning paths by providing fine-grained, step-by-step verification signals that SFT fails to capture.
Notably, while RLVR delivers substantial gains on these base models, the improvements on heavily distilled reasoning models like DeepSeek-R1-0528-Qwen3-8B are more modest (e.g., $0.579$ to $0.608$ in Problem Solving). 
This phenomenon is intuitive: models like DeepSeek-R1-0528 are already extensively optimized through large-scale reasoning distillation and reinforcement learning, resulting in highly structured reasoning trajectories with limited margin for reward reshaping. 
Nevertheless, RLVR still provides consistent improvements and systematically reduces physical and logical violations even in this heavily distilled setting. 
This highlights that VRM-based supervision is highly effective for base models with weaker domain-aligned reasoning, while serving as a complementary and rigorous refinement step for already highly aligned models.

\paragraph{Parameter Efficiency \& Proprietary Models.}
Our approach demonstrates that precise reward engineering can compensate for model scale. The VRM-optimized Qwen3-8B ($0.680$ in Problem Solving) outperforms significantly larger models, including ChatGPT-5 ($0.642$) and DeepSeek-R1-0528 ($0.665$). This suggests that open-weights models, when optimized with high-quality verifiable rewards, can bridge the gap with frontier proprietary systems.

\paragraph{Inference-Time Scaling and Distillation.}
We evaluate the discriminative power of our VRM using Best-of-$N$ (BoN) sampling. As shown in Fig.~\ref{fig:bon_scaling}, performance scales monotonically with sample size $N$, validating the VRM's effectiveness as a verifier. Notably, our RLVR policy (single sample) matches the performance of the SFT policy with BoN ($N=32$). This confirms that RLVR effectively distills the verification signal into the policy parameters, achieving the accuracy of extensive inference-time search with significantly reduced computational cost.

\paragraph{Comparison with Human Experts.}
Despite these improvements, a gap remains between the best model and human experts. Human experts maintain a clear advantage in precision-oriented tasks (Fill-in-the-blank, True/False), attributed to superior lexical exactness and robustness against conceptual nuances. Conversely, models outperform humans in complex generative tasks (Problem Solving), leveraging their vast computational capacity and knowledge synthesis. This indicates that while models excel at broad reasoning, they still lack the rigorous discrimination required for expert-level precision.

\subsection{Generalization to Alternative RL Algorithms}
\label{sec:grpo_generalization}

To further demonstrate the algorithm-agnostic nature of our VRM, we evaluated its integration with GRPO. As shown in Table~\ref{tab:model_comparison_final}, integrating VRM with GRPO on Qwen3-8B yields consistent improvements over the standard SFT baseline across all five evaluation metrics. This demonstrates that our VRM is not restricted to PPO, but is also highly compatible with group-based policy optimization. Furthermore, while the GRPO variant slightly underperforms our fully-tuned PPO baseline, a gap we attribute to the restricted sampling budget, it exhibits a similar trajectory of improvement over SFT. Overall, these findings indicate that VRM is a robust and adaptable mechanism for enhancing verifiable reasoning across different reinforcement learning paradigms.

\subsection{Generalization on Scientific Benchmarks}
\label{subsec:generalization}
We assess out-of-distribution (OOD) generalization on standard scientific benchmarks (Table~\ref{tab:scientific_generalization}). Our RLVR method consistently surpasses the SFT baseline, boosting accuracy on the quantum subsets of both \textsc{SuperGPQA} and \textsc{Physics}.
These gains indicate that our verification signals foster robust reasoning capabilities beyond the source domain, effectively mitigating overfitting risks.

\begin{table}[t]
    \centering
    \small
    \begin{tabular}{lcc} 
        \toprule
        \multirow{2}{*}{\textbf{Method}} & \textbf{SuperGPQA} & \textbf{Physics} \\ 
        \cmidrule(lr){2-2} \cmidrule(lr){3-3} 
         & Acc. $\uparrow$  & Acc. $\uparrow$ \\ 
        \midrule
        \multicolumn{3}{l}{\textit{Qwen3-8B}} \\ 
        \hspace{1em} Baseline & 0.388 & 0.254 \\
        \hspace{1em} SFT       & 0.418 & 0.352 \\
        \hspace{1em} RLVR (Ours) & \textbf{0.466}  & \textbf{0.423} \\
        \bottomrule
    \end{tabular}
    \caption{Zero-shot performance on broad scientific benchmarks.}
    \label{tab:scientific_generalization}
\end{table}

\subsection{Verbosity and Efficiency Trade-off}
\label{subsec:analysis_verbosity}

\begin{figure}[t]
    \centering
    \includegraphics[width=\linewidth]{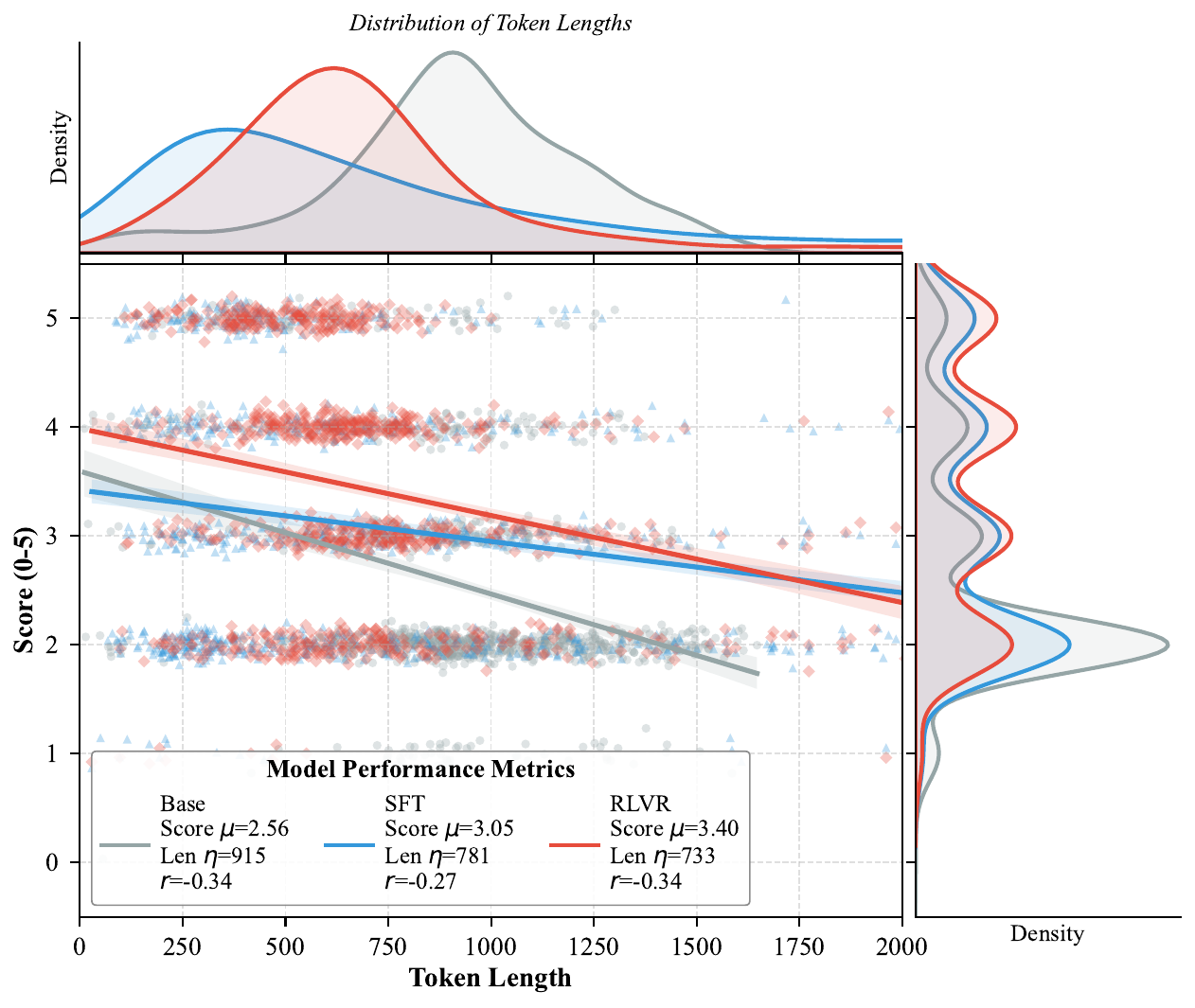} 
    \caption{Joint distribution of token length and solution quality.
    Scatter plot with marginal density estimations for Base, SFT, and RLVR models on the Qwen3-8B backbone. Solid lines represent linear regression fits. The inset table reports the statistical metrics: $\mu$ denotes the mean value for \textit{Score} (solution quality) and \textit{Len} ($\eta$, token length), while $r$ represents the Pearson correlation coefficient between length and quality.
}
    \label{fig:verbosity_analysis}
\end{figure}

To rule out reward hacking, specifically verbosity bias~\citep{singhal2024a}, we analyze the correlation between solution length and quality on Qwen3-8B (Fig.~\ref{fig:verbosity_analysis}). 
Unlike Base and SFT models which exhibit redundancy, RLVR achieves superior accuracy with the lowest token usage. 
The observed negative correlation confirms that VRM mitigates reasoning loops: by integrating verification signals, the reward is incentivized to pursue efficient derivation paths rather than exploiting sequence length.

\begin{table*}[htbp]
\centering
\small
\setlength{\tabcolsep}{5pt} 
\begin{tabular}{lcccccc}
\toprule
\multirow{2}{*}{\textbf{Method}} & \textbf{Fill-in-the-blank} & \textbf{True/False} & \textbf{Multiple Choice} & \textbf{Problem Solving} & \textbf{Short Answer} \\
\cmidrule(lr){2-2} \cmidrule(lr){3-3} \cmidrule(lr){4-4} \cmidrule(lr){5-5} \cmidrule(lr){6-6}
& Exact Match $\uparrow$ & Accuracy $\uparrow$ & Accuracy $\uparrow$ & $\mathrm{ACC}_{\mathrm{U}}$ $\uparrow$ & $\mathrm{ACC}_{\mathrm{U}}$ $\uparrow$ \\
\midrule
\textbf{RLVR (Ours)} & \textbf{0.578} & \textbf{0.912} & \textbf{0.978} & \textbf{0.680} & \textbf{0.897} \\
\midrule
\multicolumn{6}{l}{\textit{Ablation on Reward Signals}} \\
\hspace{1em} w/o mathematical correctness & 0.476 & 0.876 & 0.960 & 0.641 & 0.859 \\
\hspace{1em} w/o physical consistency & 0.560 & 0.908 & 0.971 & 0.646 & 0.872 \\
\hspace{1em} w/o instruct following & 0.566 & 0.909 & 0.971 & 0.654 & 0.892 \\
\hspace{1em} w/o math verifier & 0.549 & 0.908 & 0.968 & 0.636 & 0.877 \\
\hspace{1em} w/o physics verifier & 0.563 & 0.909 & 0.975 & 0.640 & 0.865 \\
\midrule
\multicolumn{6}{l}{\textit{Ablation on Weighting Strategies}} \\
\hspace{1em} Fixed Weight Strategy & 0.562 & 0.907 & 0.971 & 0.645 & 0.877 \\
\hspace{1em} Heuristic Weight Strategy & 0.567 & 0.912 & 0.972 & 0.648 & 0.894 \\
\midrule
\multicolumn{6}{l}{\textit{Alternative Reward Model}} \\
\hspace{1em} Skywork RM & 0.451 & 0.893 & 0.961 & 0.600 & 0.821 \\
\bottomrule
\end{tabular}
\caption{Ablation study of reward components and weighting strategies on Qwen3-8B. The best results are highlighted in \textbf{bold}.}
\label{tab:ablation_rlvr}
\end{table*}

\subsection{Fine-Grained Error Analysis}
\label{subsec:error_analysis}

\begin{figure}[t]
    \centering
    \includegraphics[width=0.9\linewidth]{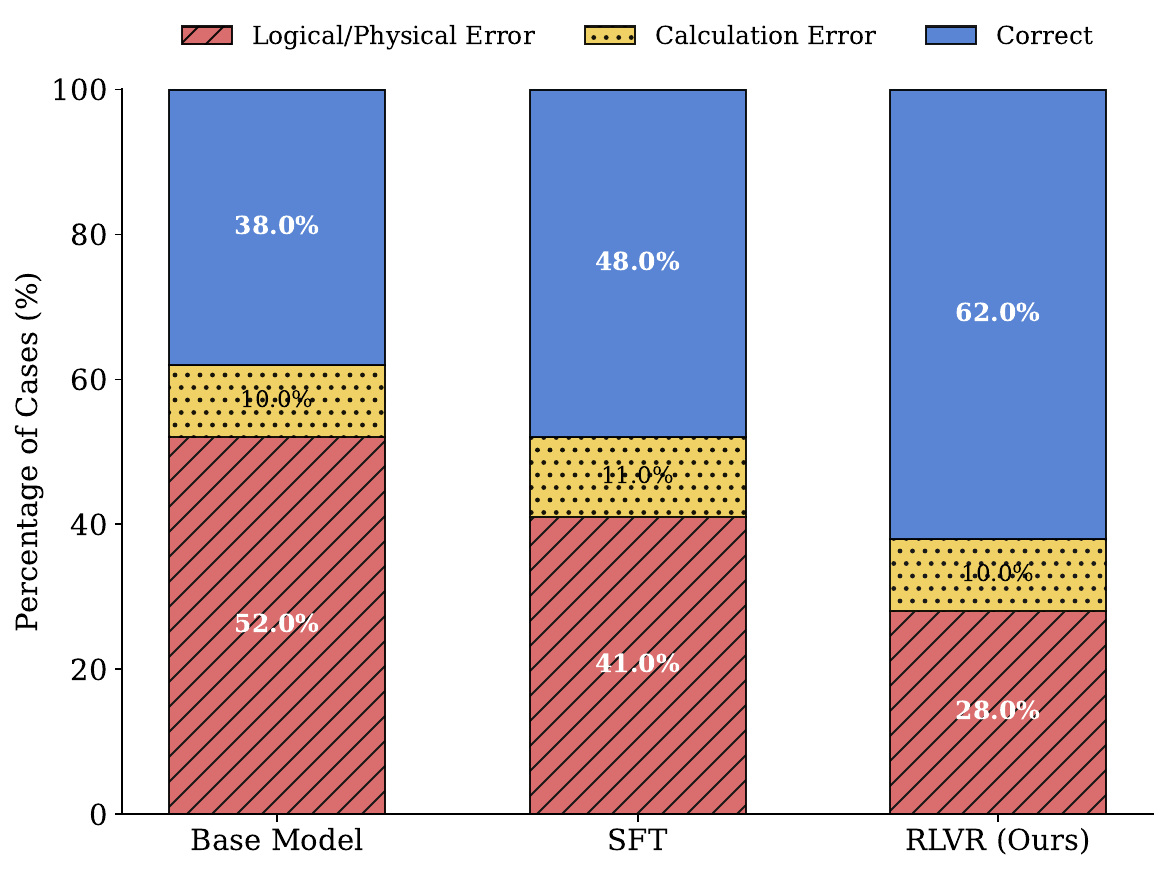} 
    \caption{
    Distribution of error types across training stages.
    }
    \label{fig:error_breakdown}
\end{figure}

To investigate the source of performance gains, we conducted a blind human evaluation on 100 sampled Problem Solving tasks, classifying errors into logical/physical violations and calculation errors. As shown in Fig.~\ref{fig:error_breakdown}, SFT exhibits a high rate of logical violations (41.0\%) with 48.0\% accuracy. In contrast, RLVR reduces logical errors to 28.0\% and improves accuracy to 62.0\%. This demonstrates that VRM acts as an effective regularizer, significantly enhancing physical consistency.

\subsection{Ablation Study}
\label{subsec:ablation}
To investigate the contribution of the verification signals and the adaptive weighting mechanism, we conduct a series of ablation experiments on Qwen3-8B.
Results are summarized in Table~\ref{tab:ablation_rlvr}.

\paragraph{Impact of Individual Verification Signals.}

We first isolate the contributions of distinct verification components.
The consistent performance degradation observed in the ablated variants validates the architectural necessity of each component. 
Additionally, ablating domain-specific verifiers (Math and Physics) leads to significant performance degradation compared to removing semantic constraints (Instruction Following). specifically, removing the math verifier causes a sharper drop ($0.680 \rightarrow 0.636$) than ablating mathematical correctness ($0.641$), indicating that deterministic feedback from external tools provides stronger supervision than internal semantic consistency. Conversely, the marginal impact of instruction ablation suggests that performance on scientific tasks is predominantly driven by rigorous reasoning signals rather than format adherence.

\paragraph{Effect of reward aggregation strategies.}
We further evaluate the ARF mechanism by comparing it with two baselines: fixed uniform weights, and heuristic task-dependent weights manually assigned based on question type. 
Both static strategies underperform the ARF, particularly on Problem Solving and Short Answer questions. 
This suggests that static or manually designed weighting schemes fail to accommodate the varying reliability of verification signals in heterogeneous queries.
In contrast, ARF effectively modulates rewards based on signal availability, attenuating noise from inapplicable verifiers while preserving informative gradients for robust and stable optimization.

\paragraph{Comparison with Alternative Reward Models.}
To verify the effectiveness of our proposed reward model, we replace the VRM with a state-of-the-art general-purpose preference model (Skywork RM)~\citep{liu2025skywork}. 
A substantial performance drop is observed across all task categories. 
This result highlights a key limitation of generic reward models in scientific domains: preference-based scoring lacks the granularity to detect subtle calculation errors or violations of physical consistency. 

\section{Conclusion}

In this paper, we addressed the critical challenges of hallucination and the lack of verifiable feedback in applying LLMs to rigorous scientific domains. 
We construct and release \textsc{QuantumQA}, a large-scale dataset that serves as a reliable resource for training and evaluation.
Building on this, we propose the VRM, a specialized feedback mechanism tailored for RLVR that dynamically integrates deterministic solvers with multidimensional semantic evaluation. 
This mechanism effectively mitigates the reward sparsity of rigid execution environments while strictly enforcing scientific rigor, thereby significantly enhancing the model's capability for reliable scientific reasoning.

\section{Limitations}

This work has several limitations that help clarify the scope of its contributions.

First, the current iteration of \textsc{QuantumQA} targets textual and symbolic derivations. 
While sufficient for verifying mathematical correctness and physical consistency, quantum mechanics often involves multimodal representations (e.g., circuit diagrams), which are not yet integrated into our synthesis pipeline.

Second, our current dataset predominantly captures formalized, post-hoc derivations, omitting the exploratory reasoning processes (e.g., trial-and-error, self-correction) central to recent Long Chain-of-Thought (Long-CoT) paradigms. 
However, this limitation motivates a highly promising direction for future research by leveraging our Problem Solving subset. 
Because these instances feature multi-step theoretical proofs (averaging over 5,000 characters) and are rigorously verified by our hybrid verification framework to prevent hallucinations, they serve as pristine prompt seeds.
Building on this, our critical next step is synthetic augmentation. We will use these seeds to elicit extended reasoning traces from advanced LLMs, subsequently filtering the outputs against our ground truth to advance Long-CoT capabilities in complex scientific domains.

Finally, regarding cross-domain generalization, our current empirical scope is limited to quantum mechanics. We selected this field as a representative, high-complexity testbed due to its rigorous axiomatic foundation and the availability of deterministic verifiers. However, it is important to note that the VRM architecture and the SES interface are designed to be domain-agnostic. The core mechanism,  integrating semantic reward signals with execution feedback, can be extended to other natural science domains (e.g., chemistry or classical mechanics) by replacing the symbolic solvers and rule checkers within the SES with corresponding domain-specific tools. Broadening our empirical validation to a wider spectrum of scientific reasoning tasks to demonstrate the framework's generalizability remains an important direction for future work.

\section*{Ethical Considerations}

\paragraph{Ethics and Data Policy.}
\textsc{QuantumQA} integrates model-generated content with publicly available materials, all utilized in compliance with their respective licenses, and will be released under a permissive open-source license. 
To ensure data privacy and quality, our pipeline employs automated filtering to remove personally identifiable information (PII) and incorporates an expert verification stage, detailed in Appendix~\ref{sec:appendix_expert_verification}, to screen for scientific errors and potential biases. 
Furthermore, this verification process received Institutional Review Board (IRB) approval, and all participating experts provided informed consent. Participants were compensated at a fair rate of \$20 USD/hour, which exceeds the local living wage. Finally, only anonymized data will be released for research purposes.

\paragraph{Intended use and risks.}
This work aims to improve the reliability of large language models in quantum mechanics using verifiable, rule-based reward signals.
While the proposed multi-dimensional reward model enhances mathematical correctness and physical consistency, it may still produce imperfect evaluations due to limitations of verification tools or learned components.
Users should avoid misuse such as reward hacking and over-reliance on automated judgments in high-stakes scientific settings, and are encouraged to perform independent verification when necessary. 

\paragraph{AI assistance.}
We used large language models, including ChatGPT, to assist with language polishing and presentation refinement, while all technical design, experiments, and conclusions were determined by the authors.

\bibliography{custom,anthology-1,anthology-2}

\appendix

\section{Related Works}
\paragraph{Scientific Question Answering Datasets}
Advancing scientific LLMs demands high-quality corpora for both effective instruction tuning and rigorous evaluation. While general benchmarks like MMLU~\citep{hendrycks2021measuring} and GPQA~\citep{rein2024gpqa} establish fundamental evaluation baselines, they are predominantly limited to multiple-choice formats and lack the granular reasoning steps necessary for training models in specialized physics. 
Consequently, focus has shifted to quantum-specific resources, yet existing works exhibit a critical trade-off between scale and scientific rigor.
On one hand, evaluation-centric datasets like QuantumBench~\citep{minami2025quantumbenchbenchmarkquantumproblem} ensure validity through expert annotation but are insufficient for training due to limited scale.
On the other, large-scale training corpora like QuantumLLMInstruct~\citep{kashani2024quantumllminstruct500kllminstructiontuning} offer extensive coverage but lack rigorous physical verification mechanisms to prevent hallucination.
Additionally, several recent datasets target quantum algorithm implementation and code generation~\citep{yang2025qcircuitbench, vishwakarma2024qiskithumanevalevaluationbenchmark, paltenghi2024surveytestinganalysisquantum}, as well as low-level circuit design and compilation tasks~\citep{foder2024reinforcementlearningvariationalquantum, fu2025qagentllmbasedmultiagentautonomous, zhang2024scalablequantumdynamicscompilation}. 
However, these resources primarily emphasize programming and circuit-level objectives, and they still do not provide scalable, process-supervised data needed to systematically elicit mathematical reasoning in quantum theory~\citep{liu-etal-2024-mathbench, vishwakarma2024qiskithumanevalevaluationbenchmark}. 
Our \textsc{QuantumQA} bridges this gap by providing a verifiable, large-scale dataset that supports both reliable training and rigorous evaluation under physics-consistent constraints.

\paragraph{Verifiable Reasoning and Alignment}
Recent advancements in mathematical reasoning have evolved from simple pre-training~\citep{lewkowycz2022solving} to sophisticated reinforcement learning pipelines that leverage CoT and process supervision~\citep{wei2022chain, shao2025deepseekmath, luo2025wizardmath}. 
However, standard alignment techniques, primarily RLHF, often fail in rigorous scientific domains. 
They tend to optimize for subjective plausibility rather than objective correctness, suffering from reward over-optimization~\citep{gao2023scaling, skalse2022defining, miao2025informationtheoreticrewardmodelingstable} and frequently yielding plausible-sounding hallucinations that violate fundamental physical constraints~\citep{ouyang2022training, taylor2022galacticalargelanguagemodel, chua2025learning}.
To address these limitations, the field has shifted toward RLVR, which grounds reasoning in execution signals~\citep{gou2024tora, gou2024critic}. 
Despite this progress, existing RLVR frameworks typically rely on sparse, outcome-based supervision~\citep{lightman2023let, uesato2022solving} manifesting as monolithic binary rewards~\citep{wu2023fine}. 
Such coarse feedback is inadequate for complex fields like quantum mechanics, which demands the satisfaction of orthogonal constraints.
While PRMs succeed in math~\citep{setlur2025rewarding, yang2025errorprocessrewardmodels} and coding~\citep{li-etal-2025-codeprm, zhang2025dreamprmcodefunctionasstepprocessreward} via dense executable signals, they face challenges in science. 
Unlike formal domains, scientific problems often lack discrete intermediate ground truths, making it difficult to define the unambiguous, step-wise checks required for reliable process supervision.
While concurrent works like TRACE~\citep{imani-etal-2026-trace} share our goal of emphasizing stepwise reasoning analysis over outcome-only supervision, they focus on Vision-Language Models, leaving an unaddressed need for physics-grounded LLMs that utilize deterministic and semantic evaluation to validate scientific trajectories.
Furthermore, existing multi-objective RL frameworks mainly prioritize balancing general-purpose human preferences, such as helpfulness and safety~\citep{wang-etal-2024-interpretable, bai2022training, dai2024safe}, while often neglecting domain-specific constraints like physical consistency. 
Moreover, these methods typically rely on static scalarization~\citep{wang2024helpsteer}, lacking adaptive reward mechanisms to accommodate the varying availability of verification signals in scientific reasoning. 
Our VRM bridges this gap by dynamically fusing deterministic symbolic verification with probabilistic semantic evaluation, adapting supervision strength to the specific verifiability of each query.

\section{Validation of LLM-as-a-Judge Evaluation}
\label{app:judge_validation}

To ensure the reliability of using an LLM-as-a-judge for our tasks, we conduct a comprehensive validation study. This study focuses on two key aspects: alignment with human expert consensus and inter-model robustness across different state-of-the-art LLMs.

\paragraph{Alignment with Human Experts}
We evaluate the consistency between our primary LLM judge and human experts on a stratified sample of 100 responses. Each response is graded by four experts on a 1--5 Likert scale, with the average score serving as the ground truth. 

As shown in Table~\ref{tab:judge_consistency}, the human inter-annotator agreement is substantial (ICC = 0.80), providing a reliable baseline. The LLM judge demonstrates robust alignment with human consensus, achieving an overall Spearman's $\rho$ of 0.82. Notably, Problem Solving tasks exhibit stronger alignment ($\rho=0.84$) compared to Short Answer tasks($\rho=0.79$). We attribute this to the fact that mathematical and scientific derivations follow rigid logical rules, whereas Short Answer questions often involve semantic nuances that introduce greater subjectivity. Although a slight calibration offset exists (MAE = 0.55), the high rank correlation confirms the model's reliability for relative comparative evaluation, which is the primary focus of our experiments.

\begin{table}[htbp]
\centering
\footnotesize
\begin{tabular}{lccc} 
\toprule
\textbf{Task Type} & \textbf{QWK} & \textbf{Spearman $\rho$} & \textbf{MAE} \\
\midrule
Overall & 0.78 & 0.82 & 0.55 \\
\midrule
Problem Solving & 0.81 & 0.84 & 0.48 \\
Short Answer & 0.75 & 0.79 & 0.62 \\
\bottomrule
\end{tabular}
\caption{Agreement between LLM judge and human experts. Human inter-annotator agreement (ICC) is 0.80. The model shows strong ranking capability (High $\rho$) despite absolute scoring deviation (MAE).}
\label{tab:judge_consistency}
\end{table}

\paragraph{Robustness Across Different LLM Judges}
To ensure our evaluation metric is not biased toward specific model artifacts from Qwen3-Max, we further compare its judgments against other highly capable models, specifically Kimi-K2-Thinking and ChatGPT-5. All models were evaluated using identical prompts and parameters to ensure comparability.

\begin{table}[htbp]
\centering
\footnotesize
\begin{tabular}{lcc} 
\toprule
\textbf{Judge Model} & \textbf{Spearman $\rho$} & \textbf{Pearson $r$} \\
 & (vs. Human) & (vs. Qwen3-Max) \\
\midrule
Qwen3-Max (Ours) & 0.82 & - \\
Kimi-K2-Thinking & 0.76 & 0.85 \\
ChatGPT-5 & 0.83 & 0.87 \\
\bottomrule
\end{tabular}
\caption{Consistency across different LLM judges. All evaluated models demonstrate high validity against human experts and strong inter-model agreement, indicating a shared capability in capturing output quality.}
\label{tab:inter_model_agreement}
\end{table}

As presented in Table~\ref{tab:inter_model_agreement}, all evaluated judges demonstrate high validity, exhibiting strong rank correlations with human experts ($\rho \ge 0.76$) and high inter-model agreement. These results suggest that despite architectural differences, these LLMs consistently capture similar underlying dimensions of output quality. Given these quantitative findings, we conclude that the specific judge used in our main experiments (Qwen3-Max) provides a fair, robust, and unbiased evaluation for this scientific reasoning task.

\section{Experimental Settings}
\label{sec:appendix_implementation}

We provide the comprehensive experimental settings for the our training pipeline: SFT, VRM training, and RLVR. 
All models are trained using bfloat16 precision to optimize memory efficiency without compromising numerical stability.

\subsection{Hyperparameter Configuration and Search}
\label{subsec:hyperparameter_search}

Given the computational constraints of large language model training, we adopt established conventions from recent alignment literature~\citep{ouyang2022training, rafailov2023direct} rather than performing an exhaustive grid search. We conduct a targeted search for the most sensitive hyperparameters, specifically the learning rate and batch size.

\paragraph{Selection Process.}
For the SFT and VRM stages, we sweep over learning rates $\alpha \in \{\num{5e-6}, \num{1e-5}, \num{2e-5}\}$. We select $\alpha = \num{1e-5}$ for SFT and $\alpha = \num{5e-6}$ for VRM training, as these values achieve the lowest validation loss without overfitting.

For the RLVR stage, optimization stability is paramount. We find that standard learning rates often lead to policy collapse. Consequently, we search for a conservative learning rate in the range $\{\num{1e-7}, \num{3e-7}, \num{5e-7}\}$. We identify that a lower learning rate of \num{3.0e-7} provides the most stable improvement in reward signals while maintaining KL divergence within a reasonable range. The best-found hyperparameters are reported in Table~\ref{tab:hyperparameters}.

\subsection{Supervised Fine-Tuning}
\label{subsec:sft_details}

Before reinforcement learning, we perform SFT to initialize the policy model. This stage ensures the model acquires the foundational capabilities required for instruction following and format compliance. 
We train the model for $3$ epochs with a global batch size of 128. We utilize a cosine learning rate scheduler with a peak learning rate of \num{1.0e-5} and a warmup ratio of $0.1$. This conservative schedule prevents catastrophic forgetting of pre-trained knowledge while adapting to scientific reasoning tasks.

\subsection{VRM Training}
\label{subsec:vrm_training_details}

To establish a robust reward signal, we train the VRM prior to the RL stage. Utilizing the Qwen3-4B backbone, we initialize the model weights and attach the randomly initialized Scoring and DWA heads. The model is trained on the Oracle-annotated dataset $\mathcal{D}$ (described in \S\ref{subsec:oracle_training}) to minimize the multi-task regression loss.

We employ a global batch size of 64 and train for 4 epochs. To ensure precise regression of the soft scores and reliability weights, we utilize a reduced learning rate of \num{5.0e-6} with a linear decay schedule. The resulting model is then frozen and used as the critic during the subsequent RLVR phase.

\subsection{RLVR with Verification-Aware Reward Model}
\label{subsec:rlvr_details}

In the final phase, we employ reinforcement learning to optimize the SFT-initialized policy model against the scalar reward provided by the VRM. While our primary experiments utilize PPO, we also evaluate our framework using GRPO to demonstrate its algorithm-agnostic nature.

\paragraph{Generation and Rollout.}
In this phase, we generate responses with a maximum length of 4096 tokens to accommodate complex reasoning chains. To encourage diverse exploration of the solution space, we employ nucleus sampling with $p=0.85$ and top-$k$ sampling with $k=50$, setting the temperature to $0.6$.

\paragraph{PPO Optimization.}
For our primary optimization phase using PPO, we set the learning rate to \num{3.0e-7}, which is significantly lower than that of the SFT phase to ensure stable policy updates. The PPO algorithm runs for 4 epochs per batch with a buffer size of 1.

\paragraph{Generalization to GRPO.}
In this configuration, we maintain the learning rate at \num{3.0e-7} and set the group size to 16. The effective batch size is configured based on a per-device count of 5 with 3 gradient accumulation steps.

\subsection{Computational Budget}
All training stages were conducted on a single compute node equipped with 8 NVIDIA H200 GPUs using full 8-way parallelism. We report the approximate wall-clock time and the corresponding total GPU hours for each phase as follows:
\begin{itemize}
    \item SFT Phase: $\approx$ 5 hours (40 GPU hours) per model.
    \item VRM Phase: $\approx$ 10 hours (80 GPU hours) per model.
    \item RLVR Phase: $\approx$ 8 hours (64 GPU hours) per model.
\end{itemize}

\subsection{Software and Implementation Details}
Our implementation is built upon the PyTorch framework~\citep{paszke2019pytorchimperativestylehighperformance} (v2.1.2) and the Hugging Face Transformers library~\citep{wolf2020huggingfacestransformersstateoftheartnatural} (v4.37.0). 
For the reinforcement learning (PPO) phase, we utilize the TRL library~\citep{havrilla-etal-2023-trlx} with standard hyperparameter configurations unless otherwise specified. 
Regarding evaluation, to ensure reproducibility and consistency with prior baselines, we use the official evaluation scripts provided by the respective benchmark creators without modification. 
For any text generation metrics (if applicable), we employ the Hugging Face evaluate library (v0.4.0) with default tokenizer settings.

\begin{table}[htbp]
    \centering    
    \setlength{\tabcolsep}{3.5pt} 
    \resizebox{\columnwidth}{!}{ 
    \begin{tabular}{lccc}
    \toprule
    \textbf{Hyperparameter} & \textbf{SFT} & \textbf{VRM} & \textbf{RLVR (PPO)} \\
    \midrule
    Backbone & Qwen3-8B & Qwen3-4B & Qwen3-8B \\
    Per-device Batch Size & 8 & 4 & 5 \\
    Gradient Accumulation & 2 & 2 & 2 \\
    Global Batch Size & 128 & 64 & 80 \\
    Learning Rate & \num{1.0e-5} & \num{5.0e-6} & \num{3.0e-7} \\
    LR Scheduler & Cosine & Linear & Cosine \\
    Warmup Ratio & 0.1 & 0.03 & 0.1 \\
    Num. Epochs & 3 & 4 & 1 \\
    \midrule
    \textit{Specific Configs} & & & \\
    Loss Function & Cross-Ent. & MSE & PPO-Clip \\
    Max Tokens & 4096 & 2048 & 4096 \\
    Temperature & - & - & 0.6 \\
    Top-$k$ / Top-$p$ & - & - & 50 / 0.85 \\
    PPO Epochs & - & - & 4 \\
    \bottomrule
    \end{tabular}
    }
    \caption{Hyperparameters used for SFT, VRM training, and RLVR (PPO). The VRM serves as a frozen reward signal during the RLVR stage.}
    \label{tab:hyperparameters}
\end{table}

\section{Detailed Dataset Construction and Verification Pipeline}
\label{sec:appendix_pipeline}

To ensure procedural transparency and facilitate reproducibility, we detail the specific implementation of our data synthesis pipeline.
As illustrated in Fig.~\ref{fig:data_construct}, the construction pipeline operates as an iterative, two-phase protocol designed to transform authoritative knowledge into high-quality instruction data while ensuring rigorous physical fidelity.

\subsection{Phase 1: Task-Adaptive Data Construction}
This phase employs a strictly ordered generation pipeline to derive instruction data from foundational texts.

\paragraph{Step 1: Seed Initialization (Source $\rightarrow$ Seed)}
We first digitize authoritative quantum mechanics and quantum information textbooks~\citep{griffiths2018introduction, nielsen2010quantum} via Optical Character Recognition. Subsequently, we prompt DeepSeek-V3 to extract core theorems and principles from the processed text. 
To eliminate redundancy, we utilize a pre-trained sentence encoder (all-MiniLM-L6-v2) to evaluate semantic similarity, filtering out entries that exceed a cosine similarity threshold of $\Upsilon = 0.85$.
This ensures a unique and high-quality seed repository.

\paragraph{Step 2: Hierarchical Concept Decomposition (Seed $\rightarrow$ Topic)}
To ensure granular conceptual coverage, we leverage Qwen3-Max, guided by specific system prompts, to systematically decompose each validated seed into a diverse array of fine-grained sub-topics. We again employ the semantic deduplication protocol on the generated outputs, rigorously filtering out redundancies to maintain a highly diverse topic pool.

\paragraph{Step 3: Data Generation (Topic $\rightarrow$ Question/Answer Pair)}
To maximize dataset diversity, we adopt a heterogeneous model ensemble generation strategy comprising DeepSeek-V3, Qwen3-Max, and ChatGPT-5. Tasks are bifurcated into deterministic and nondeterministic streams. Global dataset diversity is maintained via the aforementioned semantic similarity threshold ($\Upsilon = 0.85$).

\begin{itemize}
    \item Deterministic Tasks: For closed-ended formats, we employ an end-to-end generation process. The ensemble generates the question stem ($Q$), the ground-truth answer ($A$), distractors (if applicable), and a difficulty label ($L \in \{\text{Undergraduate}, \text{Graduate}, \text{Research}\}$).
    \begin{itemize}
        \item Multiple Choice: The model generates $Q$, the correct option ($A_{\text{correct}}$), and three plausible distractors ($A_{\text{distractors}}$) based on common quantum mechanics misconceptions to ensure high discriminative power. Outputs are constrained to option letters.
        \item True/False: The model generate declarative statements requiring deep conceptual understanding to verify. Outputs are strictly boolean.
        \item Fill-in-the-Blank: The model selects a key theorem or definition and masks critical variables, ensuring the provided context is sufficient for a unique, deterministic completion. The output contains only the missing term.
    \end{itemize}

    \item Nondeterministic \& Complex Tasks: For open-ended or calculation-heavy tasks, we adopt a two-phase stepwise generation pipeline.
    \begin{enumerate}
        \item Question Generation (Topic $\rightarrow$ $Q$): The ensemble generates high-complexity queries, ranging from definitional synthesis to complex, multi-step derivations seeded with high-difficulty parameters.
        \item Answer Derivation \& Profiling ($Q \rightarrow A + L$): The models produce a structured standard solution simulating a standard textbook answer key, alongside a difficulty label ($L$) that serves as a gating signal for subsequent reasoning injection.
    \end{enumerate}
\end{itemize}

\paragraph{Step 4: Adaptive CoT Injection ($L$ + $Q$ + $A \rightarrow$ Chain-of-Thought)}
We implement an adaptive mechanism to determine the necessity of explicit reasoning traces. We feed the triplet $(Q, A, L)$ into our high-capability ensemble, instructing it to autonomously evaluate the cognitive load required to bridge $Q$ and $A$.

If the model determines that the transition requires logical derivation or multi-step calculation (typically for Problem Solving tasks), it generates a rigorous, reverse-engineered derivation encapsulated within \texttt{<think>} tags. Conversely, for direct fact retrieval, tag generation is bypassed to prevent reasoning hallucination. Valid reasoning traces are parsed and explicitly formatted as the thought process in the final training dataset.

\subsection{Phase 2: Hybrid Verification Protocol}
To ensure the synthesized data meets the rigorous mathematical and physical standards required by the quantum science domain, we implement a dual-layer filtering mechanism.

\paragraph{Layer 1: Automated Verification}
Before human review, all raw synthetic samples undergo automated verification to filter out hallucinations and calculation errors.
\begin{itemize}
    \item Symbolic Executors: We utilize the SES to computationally verify mathematical correctness and physical consistency.
    \item Auxiliary Critic: A strong, independent LLM critic evaluates logic and formatting. This critic is strictly decoupled from the Phase 1 synthesis ensemble to ensure independent scrutiny and prevent self-reinforcing hallucinations.
\end{itemize}

\paragraph{Layer 2: HITL Audit}
Samples that pass Layer 1 undergo a rigorous HITL audit. We implement a stratified sampling strategy to monitor batch-level performance. If the batch error rate exceeds a strict threshold ($\tau > 5\%$), the entire batch is rejected. Crucially, error patterns are systematically analyzed to refine the Phase 1 prompt templates and trigger targeted regeneration.

\section{Construction Pipeline of the VRM Training Dataset}
\label{sec:appendix_dataset_pipeline}

The dataset $\mathcal{D} = \{(x, y, \mathbf{v}, \mathbf{s}^*, \mathbf{w}^*)\}$ is constructed through a rigorous 5-stage pipeline to ensure diversity, mitigate information leakage, and provide high-quality supervision signals.

\subsection{Dual-Source Prompt Engineering}
\begin{itemize}
    \item \textbf{In-Domain Prompts:} We utilize the same topic distribution as QuantumQA but use a separate generator (GPT-4o) to synthesize entirely new questions. To guarantee zero leakage, we perform strict semantic de-duplication using Sentence-Transformers (all-MiniLM-L6-v2). Any generated prompt with a semantic similarity $>0.85$ against the QuantumQA dataset and other existing prompts is automatically discarded.
    
    \item \textbf{General-Domain Prompts:} We interleave prompts from high-quality general reasoning datasets (\textsc{MATH}, \textsc{CommonsenseQA}, \textsc{PIQA}). These prompts also undergo the same BERT-based de-duplication process to ensure no accidental overlap.
    
    \item \textbf{Boundary \& Constraint Perturbation:} For both in-domain and general-domain prompts, we apply a boundary mutation strategy. Specifically, we prompt a language model (GPT-4o) to generate new problem statements based on existing prompts while explicitly altering boundary conditions, physical constraints, or initial assumptions. This creates a set of adjacent problems that test the model's ability to handle subtle logical shifts. All mutated prompts are subjected to the same strict BERT-based semantic filtering.
\end{itemize}

\subsection{Adversarial Response Generation}
We employ a matrix of models spanning three orders of magnitude in parameter size. This includes lightweight models (Qwen2.5-1.5B/7B, Meta-Llama-3.1-8B-Instruct) for simulating fundamental logic gaps, and frontier models (Qwen2.5-72B, Grok-3, ChatGPT-o1) for generating high-complexity reasoning traces.

For each question $x$, we adopt a dual-track prompting strategy. A standard prompt encourages correct, step-by-step reasoning to produce candidate positive responses $y^+$, while an adversarial prompt explicitly induces plausible but incorrect reasoning by imposing faulty constraints or assumptions, yielding hard negative samples $y^-$. Each model generates 3 responses per question from both tracks, resulting in a balanced mixture of correct solutions and challenging errors. 

\subsection{Deterministic Execution}
All generated code traces are executed via our SES to obtain a hard verification label $\mathbf{v}$. This step provides a ground truth anchor for the subsequent phase.

\subsection{Heterogeneous Oracle Annotation}
We explicitly design oracle prompts to condition on each tuple $(x, y, \mathbf{v})$, requiring the Oracle Ensemble (Anthropic Claude 3.5 Sonnet and Google Gemini 2.5 Pro) to jointly produce semantic quality assessments and adaptive weights. Concretely, the oracles are instructed to:
\begin{itemize}
    \item \textbf{Generate Soft Quality Scores ($\mathbf{s}^*$)} over multiple semantically grounded dimensions, including \textit{Corr}, \textit{Phys}, and \textit{Inst}. These scores provide fine-grained supervision beyond outcome-level correctness, enabling the VRM to internalize structured evaluation signals.
    
    \item \textbf{Assign Dimension Weights ($\mathbf{w}^*$)} conditioned on verification outcomes $\mathbf{v}$. Through prompt constraints, the oracles are encouraged to place higher weights on dimensions that are explicitly verifiable, while down-weighting other dimensions.
\end{itemize}
This design allows the VRM to learn not only how to score responses along multiple semantic axes, but also how to adaptively reweight these axes based on task and verification outcomes.

\subsection{Human-in-the-Loop Audit}
Domain experts conduct batch sampling (5\% of the data) to verify the balance of positive and negative samples and the consistency between SES execution results and Oracle scores.

\section{Implementation Details of SES}
\label{sec:appendix_ses}

This section provides an extended technical description of SES, detailing its underlying verification logic and parser-executor pipeline.

\subsection{Verification Logic and Script Composition}
The SES is implemented as a specialized library comprising 12 modular, atomic verification scripts. Each script targets a specific class of constraints to detect outputs that may appear syntactically plausible but violate underlying rigorous laws. These verifiers are categorized into two primary groups:
\begin{itemize}
    \item \textbf{Mathematical Consistency Checks:} These scripts validate symbolic equivalence and numerical correctness, ensuring the rigor of algebraic derivations and arithmetic precision.
    \item \textbf{Physical Consistency Checks:} These scripts enforce core quantum-mechanical axioms. Examples include verifying the unitarity of evolution operators and validating density matrices through positivity and trace constraints.
\end{itemize}

\subsection{The Parser--Executor Pipeline}
To reliably bridge the gap between natural language generations and deterministic code execution, the SES employs a robust, three-stage pipeline rather than relying on brittle heuristic string matching (e.g., regular expressions):
\begin{enumerate}
    \item \textbf{Semantic Parsing:} An instruction-following Large Language Model (e.g., GPT-4o) acts as a semantic parser to extract targeted variables, such as scalars, matrices, or symbolic expressions, from the model's response.
    \item \textbf{Type Casting and Parameter Passing:} The extracted elements are systematically converted into structured programmatic objects and passed as arguments to the corresponding verification script.
    \item \textbf{Execution and Exception Handling:} The targeted script executes the verification logic and returns a deterministic boolean outcome. Crucially, if the semantic parser fails to extract valid arguments due to ambiguous formatting or incomplete reasoning, the pipeline explicitly flags the sample as unparsable. This strict exception handling ensures that positive reinforcement is exclusively reserved for well-formed, verifiable derivations.
\end{enumerate}

\section{Granular Error Analysis of Verification Signals}
\label{sec:appendix_error_analysis}

To better understand the specific contributions and complementarity of the verification methods, we conducted a granular error analysis on 500 randomly sampled reasoning trajectories.

\subsection{Complementarity of Deterministic and Semantic Verification}
We first classify the sampled trajectories based on two orthogonal verification signals to analyze their overlap and divergences:
\begin{itemize}
    \item \textbf{Semantic Verification:} Driven by the semantic scoring head, which predicts a continuous score $s_p \in [0, 1]$. We apply a strict threshold $\zeta = 0.8$ to categorize a trajectory as semantically valid.
    \item \textbf{Deterministic Verification:} A stringent binary check representing the aggregate outcome of the SES. A trajectory is deemed verified only if it produces an executable format and satisfies both mathematical correctness and physical consistency constraints.
\end{itemize}

As shown in Table~\ref{tab:confusion_matrix}, a significant portion of the samples (31.6\%) failed the strict deterministic check, whether due to syntax parsing errors, minor mathematical calculation deviations, or strict violations of physical axioms, yet successfully passed the semantic verification. 
This demonstrates that the model frequently captures the correct semantic reasoning even when it struggles with precise programmatic implementation or exact constraint satisfaction. 
Our hybrid method successfully rewards these partially correct trajectories, preventing the RL agent from discarding valuable semantic insights while simultaneously mitigating the reward sparsity problem inherent in binary execution feedback. 
The semantic verifier evaluates the underlying reasoning process, whereas the deterministic verifier enforces rigorous final mathematical and physical validity.

\begin{table}[h]
\centering
\small
\begin{tabular}{@{}lccc@{}}
\toprule
 & \textbf{Passed Sem.} & \textbf{Failed Sem.} & \textbf{Total} \\ \midrule
\textbf{Passed Det.} & 166 (33.2\%) & 25 (5.0\%) & 191 (38.2\%) \\
\textbf{Failed Det.}  & 158 (31.6\%) & 151 (30.2\%) & 309 (61.8\%) \\ \midrule
\textbf{Total}        & 324 (64.8\%) & 176 (35.2\%) & 500 (100\%) \\ \bottomrule
\end{tabular}
\caption{Confusion matrix of deterministic (Det.) vs. semantic (Sem.) verification signals.}
\label{tab:confusion_matrix}
\end{table}

\subsection{Dimensional Analysis of Semantic Verification}
To further investigate the specific bottlenecks within the reasoning process, we evaluated the semantic verification across the three aforementioned dimensions: \textit{Corr}, \textit{Phys}, and \textit{Inst}.
Table~\ref{tab:semantic_evaluation} presents the pass rates for each dimension within the same 500-sample subset.

\begin{table}[h]
\centering
\small
\begin{tabular}{@{}lc@{}}
\toprule
\textbf{Dimension} & \textbf{Pass Rate} \\ \midrule
\textit{Phys} & 68.0\% \\
\textit{Corr} & 81.4\% \\
\textit{Inst} & 92.6\% \\ 
\bottomrule
\end{tabular}
\caption{Pass rates across individual semantic evaluation dimensions ($N=500$).}
\label{tab:semantic_evaluation}
\end{table}

The evaluation reveals that \textit{Phys} is the most challenging dimension, exhibiting the lowest pass rate (68.0\%). This highlights its critical role as the primary filter for plausible but fundamentally non-physical solutions. \textit{Corr} functions as an intermediate logic checker, filtering out cases where physically valid terms are combined through erroneous computations. 
Conversely, \textit{Inst} achieves a high pass rate (92.6\%), indicating that surface-level formatting constraints are learned reliably. This allows the RL optimization to concentrate on the deeper physical and mathematical semantics. By aggregating these dimensions, the semantic verifier provides dense, structured feedback that effectively guides the model through the complex landscape of quantum physics problem solving.

\section{Expert Verification}
\label{sec:appendix_expert_verification}

To ensure the correctness and reliability of \textsc{QuantumQA}, we implemented a rigorous expert verification pipeline involving 15 domain experts, comprising 10 senior Ph.D. candidates and 5 postdoctoral researchers in theoretical physics and quantum information science. Distinct from generalist crowdsourcing, this cohort was recruited specifically for their proficiency in assessing quantum derivations and validating adherence to physical axioms.

Experts served as verifiers via a custom annotation interface integrated with SES, which facilitated the on-demand verification of intermediate computational steps. The annotation guidelines standardized four evaluation metrics used for both dataset curation and model output verification: (1) correctness regarding physical principles, (2) logical coherence of the reasoning process, (3) clarity of the problem formulation, and (4) readability of the solution. Additionally, experts were instructed to explicitly flag any potential bias or inappropriate content encountered during the review.

\section{Dataset Details}
\label{sec:appendix_dataset}

\subsection{Additional Statistics}
\label{sec:appendix_stats}

In this section, we provide detailed statistics regarding the task formats, domain coverage, and difficulty distribution of \textsc{QuantumQA}.

\begin{figure}[htbp]
  \centering
  \includegraphics[width=\columnwidth]{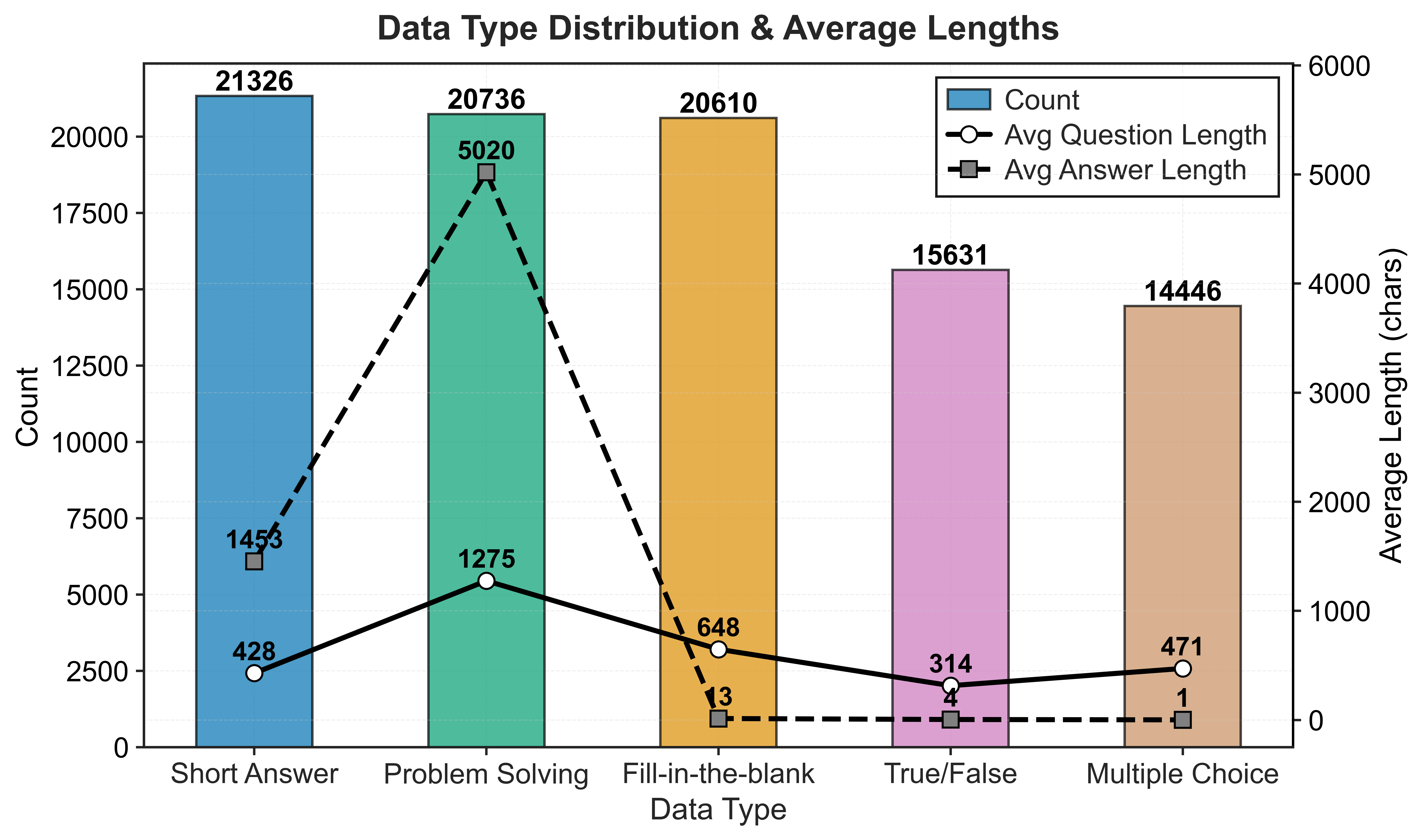}
  \caption{Distribution of question types in \textsc{QuantumQA}.}
  \label{fig:datatype_combined}
\end{figure}

\paragraph{Task Heterogeneity.}
Figure~\ref{fig:datatype_combined} illustrates the distribution of the five task formats included in the dataset: Short Answer, Fill-in-the-Blank, True/False, Multiple Choice, and Problem Solving. Notably, Problem Solving tasks constitute the majority and exhibit significantly longer average answer lengths. This distribution reflects the dataset's design philosophy, prioritizing multi-step derivation and structured mathematical reasoning over rote memorization.

\begin{figure}[htbp]
  \centering
  \includegraphics[width=\columnwidth]{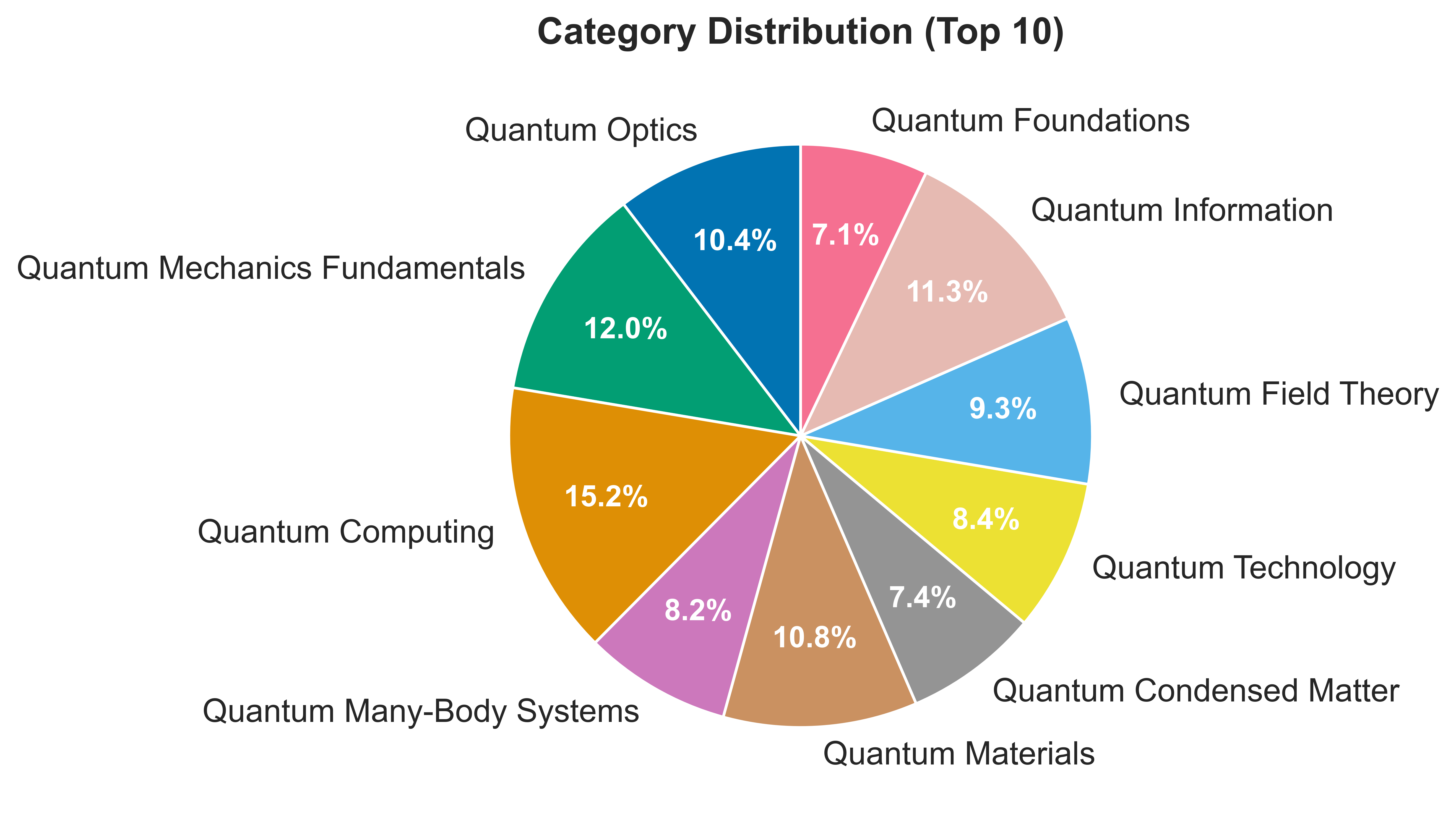}
  \caption{Top-10 category distribution of \textsc{QuantumQA}. The balanced coverage across diverse subfields ensures robust training and evaluation of quantum reasoning. Note that the cumulative percentage exceeds 100\% as samples may be annotated with multiple topic labels.}
  \label{fig:category_distribution_pie}
\end{figure}

\begin{figure}[htbp]
    \centering
    \includegraphics[width=\columnwidth]{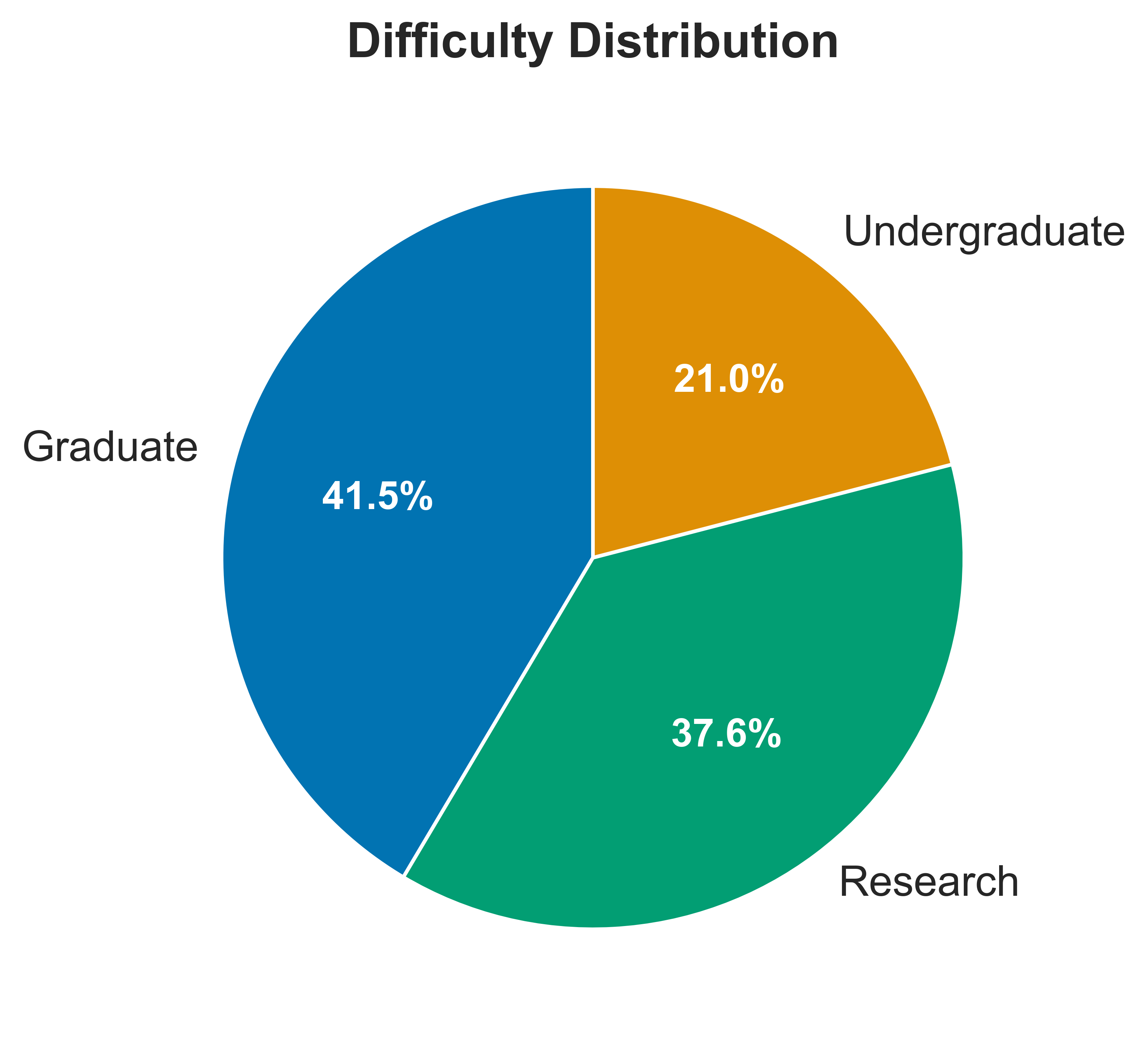} 
    \caption{Distribution of difficulty levels in \textsc{QuantumQA}. The levels are categorized based on the number of reasoning steps and the complexity of physical concepts involved.}
    \label{fig:difficulty_pie}
\end{figure}

\paragraph{Domain, Difficulty, and Language.}
To ensure robust training and evaluation, we maintain balanced coverage across diverse subfields and difficulty levels. Fig.~\ref{fig:category_distribution_pie} and Fig.~\ref{fig:difficulty_pie} depict the specific distributions for topic categories and reasoning difficulty. Regarding linguistic scope, the dataset consists exclusively of English text, covering standard scientific statement and mathematical reasoning patterns.

\subsection{Dataset Examples}
\label{sec:appendix_examples}

Table~\ref{tab:detailed_examples} provides comprehensive examples from the \textsc{QuantumQA} dataset.

\begin{table*}[htbp]
    \centering    
    \small
    \begin{tabular}{p{0.96\textwidth}}
        \toprule
        \textbf{ID:} 04215 \hfill 
        \textbf{Type:} Problem Solving \hfill 
        \textbf{Topic:} Quantum Optics \hfill 
        \textbf{Difficulty:} Hard \hfill 
        \textbf{SES Status:} Corr \textbf{\texttt{Pass}} $\mid$ Phys \texttt{Pass} \\
        \midrule
        \textbf{[System Prompt]} \\
        You are an expert quantum physicist. Solve the following problem efficiently. Ensure strict adherence to mathematical logic and physical laws. \\
          
        \textbf{[Instruction]} \\
        Calculate the time evolution of the annihilation operator in the Heisenberg picture for a single-mode cavity field and verify the commutation relation consistency. \\
          
        \textbf{[Question]} \\
        Consider a Hamiltonian $H = \hbar \omega a^\dagger a$. 
        1. Derive the expression for $a(t)$ using the Heisenberg equation of motion $\frac{d}{dt}A = \frac{i}{\hbar}[H, A]$.
        2. Prove that $[a(t), a^\dagger(t)]$ is time-independent. \\
          
        \textbf{[Answer]} \\
        \textit{Step 1: Equation of Motion} \\
        Calculate the commutator: $[H, a] = \hbar \omega [a^\dagger a, a] = \hbar \omega (a^\dagger [a, a] + [a^\dagger, a] a) = \hbar \omega (0 - a) = -\hbar \omega a$. \\
        The differential equation is: $\frac{d}{dt}a(t) = -i\omega a(t)$. \\
        \textit{Step 2: Time Evolution} \\
        Solving the ODE gives: $a(t) = a(0) e^{-i\omega t}$. Taking the Hermitian adjoint: $a^\dagger(t) = a^\dagger(0) e^{i\omega t}$. \\
        \textit{Step 3: Verification} \\
        $[a(t), a^\dagger(t)] = [a(0) e^{-i\omega t}, a^\dagger(0) e^{i\omega t}] = e^{-i\omega t} e^{i\omega t} [a(0), a^\dagger(0)] = 1 \cdot 1 = 1$. \\
        The commutator equals 1 for all $t$, confirming consistency. \\
        \midrule
        \textbf{\textit{SES Verification Log:}} \\
        \texttt{[Corr] SymPy.check(commutator\_diff == 0) -> True} \\
        \texttt{[Phys] Unitary Evolution Consistency -> Verified} \\
        \bottomrule
        \\[0.5em] 
          
        \toprule
        \textbf{ID:} 20591 \hfill 
        \textbf{Type:} Problem Solving \hfill 
        \textbf{Topic:} Bound States \hfill 
        \textbf{Difficulty:} Easy \hfill 
        \textbf{SES Status:} Corr \textbf{\texttt{Pass}} $\mid$ Phys \textbf{\texttt{Fail}} \\
        \midrule
        \textbf{[System Prompt]} \\
        You are an expert quantum physicist. Solve the following problem efficiently. Ensure strict adherence to mathematical logic and physical laws. \\
          
        \textbf{[Instruction]} \\
        Calculate the energy eigenvalue for the specified quantum state in a 1D infinite square well. \\
          
        \textbf{[Question]} \\
        A particle of mass $m$ is confined in a one-dimensional infinite potential well of width $L$. Using the standard energy formula, calculate the energy $E_n$ for the state with quantum number $n=0$. \\
          
        \textbf{[Answer]} \\
        The energy levels for a particle in a 1D infinite box are given by the formula:
        $E_n = \frac{n^2 \pi^2 \hbar^2}{2mL^2}$. \\
        Substituting the given quantum number $n=0$: \\
        $E_0 = \frac{0^2 \cdot \pi^2 \hbar^2}{2mL^2} = 0$. \\
        Thus, the energy of the particle in this state is zero. \\
        \midrule
        \textbf{\textit{SES Verification Log:}} \\
        \texttt{[Corr] SymPy.eval(subs(E\_formula, n, 0)) == 0 -> True} \\
        \texttt{[Phys] Zero-Point Energy Check -> \textbf{Failed}} \\
        \textit{Reason: Violation of Uncertainty Principle. For bound states, $n \in \mathbb{Z}^+$, minimum energy $E_1 > 0$.}  \\
        \bottomrule
    \end{tabular} 
    \caption{Detailed data samples from \textsc{QuantumQA}. Example 1 shows a successful generation. Example 2 demonstrates the framework's capability to detect physical hallucinations, where the model performs correct mathematical substitution (\textit{Corr} Pass) but violates fundamental physical constraints (\textit{Phys} Fail).} 
    \label{tab:detailed_examples}
\end{table*}

\subsection{Artifact Citations and Intended Use}
\label{subsec:artifact_citations}

We strictly adhere to the citation guidelines for the artifacts used in our experiments.
We confirm that our use of existing artifacts is consistent with their intended use, and the intended use of our created dataset is compatible with the original access conditions. Details are provided below:

\begin{itemize}
    \item We collected problem sets and definitions from standard textbooks~\citep{nielsen2010quantum, griffiths2018introduction} in the domain of Quantum Mechanics. These materials are copyrighted by their respective publishers. Our usage aligns with their intended purpose of educational assessment and domain knowledge evaluation. We operate under fair use principles, restricting usage strictly to non-commercial research contexts.
    \item We constructed the dataset using open-source Large Language Models (LLMs) in strict adherence to their respective licensing agreements. The generation of synthetic data for model alignment and evaluation is well-aligned with the intended use cases and safety guidelines specified by the model providers. This approach follows established community practices for scaling high-quality scientific reasoning data.
    \item The dataset created in this work is a derivative of the sources above. To maintain compatibility with the access conditions of the original educational materials, our dataset is intended solely for research purposes. It should not be used for commercial applications or outside of research contexts. The dataset will be released under a CC-BY-NC-4.0 license to enforce this restriction.
\end{itemize}

\section{Prompt Templates}
\label{sec:appendix_prompts}

To facilitate reproducibility and transparency, we provide the specific prompt templates used in our framework. These include the instructions for the physics-consistent data synthesis, the signal generation for our VRM, and the scalar scoring criteria.

\subsection{Prompts for Data Synthesis}
\label{subsec:prompt_data}

Table~\ref{tab:prompt_physics_math_check} presents the instruction used in our hybrid verification protocol. This prompt serves as a comprehensive filter to ensure strict adherence to physical consistency.

\begin{table*}[htbp]
    \centering
    \small
    \begin{tabularx}{\textwidth}{p{\textwidth}}
    \toprule
    \textbf{Instruction for Physical Consistency Verification} \\ 
    \midrule
    You are an expert physicist and a rigorous logic checker.\\
    Your task is to verify the physical consistency of the following quantum mechanics problem and its solution. \\
    \\
    \textbf{Input:} \\
    Problem: \{Question\} \\
    Solution: \{Solution\} \\
    \\
    \textbf{Verification Criteria:} \\
    1. Verify that the derivation respects fundamental principles (e.g., Uncertainty Principle, Commutation Relations). \\
    2. Check for dimensional homogeneity in all equations. \\
    3. Rigorously check the derivation steps, including integrals, matrix operations, and complex number arithmetic.\\
    4. Ensure each step logically follows from the previous one without gaps. \\
    \\
    \textbf{Output Format:} \\
    1. If the content is valid, output \texttt{\textbackslash boxed\{PASS\}}. \\
    2. If there are any violations, output \texttt{\textbackslash boxed\{FAIL\}} followed by a specific explanation of the error. \\
    \bottomrule
    \end{tabularx}
    \caption{The unified verification prompt used in our data synthesis framework. It enforces a rigorous standard, ensuring high-quality training data.}
    \label{tab:prompt_physics_math_check}
\end{table*}

\subsection{Prompts for Verification-Aware Reward Model}
\label{subsec:prompt_vrm}

Table~\ref{tab:prompt_vrm_generation} presents the prompt template employed to construct the VRM training data. This template integrates external verification signals (from SES) with the original query and answer to generate multidimensional soft scores and their corresponding adaptive weights, which are subsequently used to train the VRM.

\begin{table*}[htbp]
    \centering
    \small
    \begin{tabularx}{\textwidth}{p{\textwidth}}
    \toprule
    \textbf{Instruction for VRM Signal Integration and Scoring} \\
    \midrule
    \textbf{System Instruction:} \\
    You are an expert AI evaluator assessing the reasoning process of a model. Your objective is to synthesize external\\
    verifier signals with the original query and response to generate fine-grained, multidimensional soft scores and their \\
    corresponding adaptive confidence weights. \\
    \\
    \textbf{Context:} \\
    Original Query: \{Query\} \\
    Model Response: \{Response\} \\
    External Verifier Output: \{Verifier\_Signal\} (e.g., SymPy result, Execution Status) \\
    \\
    \textbf{Task:} \\
    Evaluate the response across three specific dimensions based on the query and the verifier signal. \\
    For each dimension, assign a soft score (0-1) and a confidence weight (0-1). \\
    \\
    \textbf{Dimensions:} \\
    1. \textit{Mathematical Correctness (Corr)}: Is the calculation formally correct? Use the verifier signal as ground truth. \\
    2. \textit{Physical Consistency (Phys)}: Does the reasoning follow quantum mechanics principles? \\
    3. \textit{Instruction Following (Inst)}: Did the model follow constraints (e.g., format, method)? \\
    \\
    \textbf{Output Format JSON:} \\
    \{ \\
      "scores": \{ "Corr": float, "Phys": float, "Inst": float \}, \\
      "weights": \{ "Corr": float, "Phys": float, "Inst": float \}, \\
      "rationale": "Brief explanation..." \\
    \} \\
    \bottomrule
    \end{tabularx}    
     \caption{The prompt template for the Verification-Aware Reward Model. It synthesizes verifier signals to produce fine-grained soft rewards and adaptive weights.}
    \label{tab:prompt_vrm_generation}
\end{table*}

\subsection{Prompts for Scalar Evaluation}
\label{subsec:prompt_eval}

For comparative evaluation, we employ a standard LLM-as-a-Judge prompting strategy to generate scalar scores (1-5). The template used for this assessment is shown in Table~\ref{tab:prompt_scalar_score}.

\begin{table*}[htbp]
    \centering
    \small
    \begin{tabularx}{\textwidth}{p{\textwidth}}
    \toprule
    \textbf{Instruction for Scalar Quality Scoring} \\
    \midrule
    Please rate the quality of the following answer to the quantum mechanics question on a scale from 1 to 5. \\
    \\
    \textbf{Question:} \{Question\} \\
    \textbf{Reference Answer:} \{Reference\} \\
    \textbf{Model Answer:} \{Prediction\} \\
    \\
    \textbf{Scoring Rubric:} \\
    1: Completely incorrect or irrelevant. \\
    2: Contains severe physical errors or hallucinations. \\
    3: Partially correct but contains minor logical flaws or calculation errors. \\
    4: Correct reasoning and result, but lacks clarity or detail. \\
    5: Flawless derivation, physically consistent, and clearly explained. \\
    \\
    \textbf{Output Format:} \\
    Output the integer score wrapped in a box format. For example, if the score is 5, output \texttt{\textbackslash boxed\{5\}}. \\
    \bottomrule
    \end{tabularx}    
    \caption{The prompt used for generating scalar evaluation scores (1-5).}
    \label{tab:prompt_scalar_score}
\end{table*}

\end{document}